\documentclass{article}

\usepackage{PRIMEarxiv}

\usepackage[utf8]{inputenc} 
\usepackage[T1]{fontenc}    
\usepackage{hyperref}       
\usepackage{url}            
\usepackage{booktabs}       
\usepackage{amsfonts}       
\usepackage{nicefrac}       
\usepackage{microtype}      
\usepackage{lipsum}
\usepackage{fancyhdr}       
\usepackage{graphicx}       

\usepackage{amssymb}
\usepackage{amsmath}
\usepackage{stfloats}
\usepackage{xcolor}
\usepackage{algorithm}  
\usepackage{algorithmic}
\usepackage[normalem]{ulem}
\usepackage{lscape}
\usepackage{geometry}
\usepackage{caption} 
\usepackage{longtable}
\usepackage{setspace} 
\usepackage[justification=centering]{caption}

 \usepackage{lineno}

\graphicspath{{media/}}     

\title{Improving Visual-Semantic Embeddings by Learning Semantically-Enhanced Hard Negatives for Cross-modal Information Retrieval
}

  \author{
  Yan Gong, Georgina Cosma \\
  Department of Computer Science \\
  Loughborough University \\
  \texttt{\{y.gong2, g.cosma\}@lboro.ac.uk}\\
  The paper has been accepted by Elsevier Pattern Recognition.
  }
  
\begin{document}
\maketitle

\begin{abstract}
Visual Semantic Embedding (VSE) networks aim to extract the semantics of images and their descriptions and embed them into the same latent space for cross-modal information retrieval. Most existing VSE networks are trained by adopting a hard negatives loss function which learns an objective margin between the similarity of relevant and irrelevant image--description embedding pairs. However, the objective margin in the hard negatives loss function is set as a fixed hyperparameter that ignores the semantic differences of the irrelevant image--description pairs. To address the challenge of measuring the optimal similarities between image--description pairs before obtaining the trained VSE networks, this paper presents a novel approach that comprises two main parts: (1) finds the underlying semantics of image descriptions; and (2) proposes a novel semantically-enhanced hard negatives loss function, where the learning objective is dynamically determined based on the optimal similarity scores between irrelevant image--description pairs. Extensive experiments were carried out by integrating the proposed methods into five state-of-the-art VSE networks that were applied to three benchmark datasets for cross-modal information retrieval tasks. The results revealed that the proposed methods achieved the best performance and can also be adopted by existing and future VSE networks.

\end{abstract}


\section{Introduction}\label{sec:introduction}
In information retrieval, Visual Semantic Embedding (VSE) networks aim to create joint representations of images and textual descriptions and map these in a joint embedding space (i.e. same latent space) to enable various information retrieval-related tasks, such as image–text retrieval, image captioning, and visual question answering \cite{gong2021limitations}. 
Within the shared embedding space, the aim is to position the relevant image–description pairs far away from the irrelevant pairs \cite{shu2021scalable}. 
Currently, VSE literature can be summarised into: (1) approaches that extend the cross-modal encoder-decoder network for improving the learning of latent representations crossing images and descriptions \cite{faghri2018vse++}; (2) specifically designed attention architectures that improve the performance of networks \cite{diao2021similarity}; (3) networks that are modified based on generative adversarial methods for learning the common representation of images and descriptions \cite{hu2021cross}. The above-mentioned studies aim to improve the VSE networks for information retrieval and have been evaluated using the benchmark MS-COCO \cite{lin2014microsoft} and Flickr30K~\cite{young2014image} datasets. Few studies focus on exploring the learning potential of VSE networks. The hard negatives loss function \cite{faghri2018vse++} defines the learning objective of VSE networks, and it is commonly adopted by the current VSE architectures \cite{chen2021learning}.
\begin{figure}[!ht]
\centering
	\includegraphics[width=7.5cm]{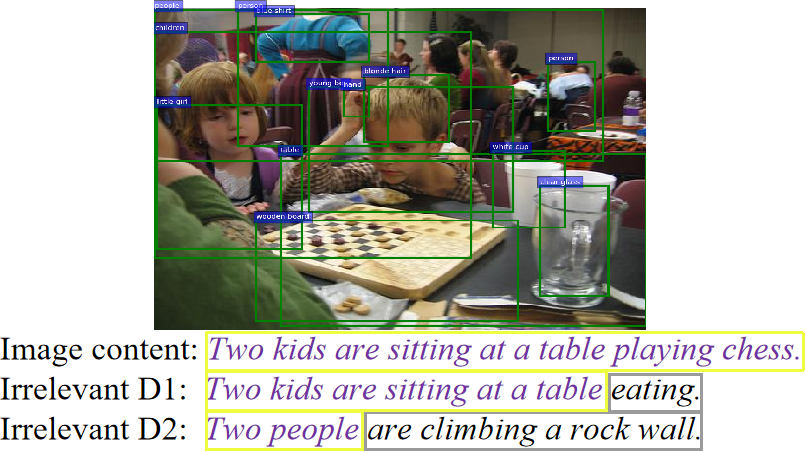}
    \caption{Sample of irrelevant image--description pairs. Description D1 is the one semantically closer to the image.}
	\label{irrExp}
\end{figure}

Furthermore, the hard negatives loss function learns a fixed margin that is the optimal difference between the similarity of the relevant image--description embedding pair and that of the irrelevant embedding pair. However, the fixed margin ignores the semantic differences between the irrelevant image–description pairs. 
The hard negatives loss function does not consider the distance of the irrelevant items to the query and sets the same learning objective (i.e. fixed margin) for both pairs, image--D1 and image--D2 (sample from Fig. \ref{irrExp}), even though the semantic differences of the irrelevant training pairs are useful for training an information retrieval model \cite{song2021deep}. To illustrate this point, consider Fig. \ref{irrExp}, where in the irrelevant image--D1 pair the image and description are semantically closer than those of the irrelevant image--D2 pair, but the hard negatives loss function sets the same learning objective for both pairs, i.e. image--D1 and image--D2, and this is not suitable. 
To solve the limitations of the fixed margin, Wei et al. \cite{wei2021universal} introduced a polynomial loss function with an adaptive objective margin, but their method does not consider the optimal semantic information from irrelevant image--description pairs. 

Our paper aims to semantically enhance the hard negatives loss function for exploring the learning potential of VSE networks. This paper (1) proposes a new loss function for improving the learning efficiency and the cross-modal information retrieval performance of VSE networks; (2) embeds the proposed loss function within state-of-the-art VSE networks, and (3) evaluates its efficiency using benchmark datasets suitable for the task of cross-modal information retrieval. The contributions of our paper are as follows. 

\begin{itemize} 
    \item A novel approach that infers the semantics of image descriptions by finding the underlying meaning of descriptions using eigendecomposition and dimensionality reduction (i.e. Singular Value Decomposition). The derived descriptions are then utilised by a proposed semantically-enhanced hard negatives loss function, entitled LSEH, when computing the optimal similarities between irrelevant image--description pairs. 
    \item A semantically-enhanced hard negatives loss function that redefines the learning objective for VSE networks. The proposed loss function dynamically adjusts the learning objective according to the semantic similarities between irrelevant image--description pairs. Ambiguous training pairs with larger optimal similarity scores obtain larger gradients that are utilised by the proposed loss function to improve training efficiency. 
    \item The proposed approach and loss function can be integrated into other VSE networks that improve learning efficiency and cross-modal information retrieval. Extensive experiments were carried out by integrating the proposed methods into five state-of-the-art VSE networks that were applied to the Flickr30K, MS-COCO, and IAPR TC12 datasets, and the results showed that the proposed methods achieved the best performance. 
\end{itemize} 

\section{Related Work}  
\textbf{VSE Networks.} VSE networks aim to align embeddings of relevant images and descriptions in the same latent space for cross-modal information retrieval \cite{gong2021limitations}. Faghri et al. \cite{faghri2018vse++} proposed an Improved Visual Semantic Embedding architecture (VSE++). Image region features extracted by the faster R-CNN \cite{anderson2018bottom} and their descriptions were embedded into the same latent space by using a fully connected neural network and a Gated Recurrent Units (GRU) network \cite{cho2014learning}. Most state-of-the-art VSE networks improve upon VSE++. Li et al. \cite{li2022image} introduced a Visual Semantic Reasoning Network (VSRN) to enhance image features with image region relationships extracted by a Graph Convolution Network (GCN)~\cite{zhang2019graph}; Liu et al. \cite{liu2020graph} applied a Graph Structured Matching Network (GSMN) to build a graph of image features and words and learn the fine-grained correspondence between image features and words; Diao et al. \cite{diao2021similarity} proposed the Similarity Graph Reasoning and Attention Filtration network (SGRAF) that extends the attention mechanisms of image and description sets. SGRAF also provides two individual sub-networks to process the attention results between the image features and the description -- where a Similarity Graph Reasoning network (SGR) builds a graph of the attention results for reasoning, and a Similarity Attention Filtration network (SAF) filters the important information from the attention results. Chen et al. \cite{chen2021learning} proposed a variation of the VSE network, VSE$\infty$, that benefits from a generalized pooling operator which discovers the best strategy for pooling image and description embeddings. 
Recently, vision transformer-based networks, that are not relying on the hard negatives loss function, have become popular for cross-modal information retrieval \cite{li2020unicoder}. However, compared to traditional VSE networks, vision transformer-based cross-modal retrieval networks require a large amount of data for training and the time they require for retrieving the results of a query makes them unsuitable for real-world applications \cite{gong2021limitations}. 
The hashing-based network is another active solution for cross-modal information retrieval \cite{duan2022ms2gah}. For example, Liu et al. \cite{liu2019mtfh} firstly proposed a hashing framework for learning varying hash codes of different lengths for the comparison between images and descriptions, and the learned modality-specific hash codes contain more semantics.  
Hashing-based networks are concerned with reducing data storage costs and improving retrieval speed. Such networks are out of scope for this paper because the focus herein is on VSE networks which mostly aim to explore the local information alignment between images and descriptions for improved retrieval performance. 

\textbf{Loss Functions for Cross-modal Information Retrieval.}  
One of the earliest and most used cross-modal information retrieval loss functions is the Sum of Hinges Loss (LSH) \cite{karpathy2015deep}. LSH is also known as a negatives loss function, and it learns a fixed margin between the similarities of the relevant image--description embedding pairs and those of the irrelevant embedding pairs. 
A more recent hard negatives loss function, the Max of Hinges Loss (LMH) \cite{faghri2018vse++}, is adopted in most recent VSE networks, due to its ability to outperform LSH \cite{liu2018learning, zhang2022multi}. An improved version of LSH, LMH only focuses on learning the hard negatives, which are the irrelevant image--description embedding pairs that are nearest to the relevant pairs. Song et al. \cite{song2021deep} presented a margin-adaptive triplet loss for the task of cross-modal information retrieval that uses a hashing-based method which embeds the image and text into a low-dimensional Hamming space. Liu et al. \cite{liu2019cyclematch} applied a variant triplet loss function into their novel VSE network for cross-modal information retrieval, where the input text embedding for the loss is replaced by the reconstructed image embedding of the network. Recently, Wei~et~al.~\cite{wei2021universal} proposed a polynomial \cite{song2019polysemous} based Universal Weighting Metric Loss (LUWM) with flexible objective margins, and that has been shown to outperform existing hard negatives loss functions.  

A summary of the limitations of the existing loss functions are as follows.   
(1) The learning objectives of LSH \cite{karpathy2015deep} and LMH \cite{faghri2018vse++} are not flexible because of their fixed margins. 
(2) The adaptive margin in \cite{song2021deep} is not optimal, because it relies on the computed similarities between irrelevant image--description embedding pairs by the training network which is optimising. 
(3) The modified ranking loss of \cite{liu2019cyclematch} cannot be integrated into other networks. 
(4) LUWM \cite{wei2021universal} does not consider the optimal semantic information from irrelevant image--description pairs. 

\textbf{VSE Networks with a Negatives Loss Function.}  LMH was proposed by Faghri et al. \cite{faghri2018vse++}, and thereafter other VSE networks adopted LMH. For improving the attention mechanism, Lee et al. \cite{lee2018stacked} proposed an approach to align the image region features with keywords of the relevant image--description pair; and Diao et al. \cite{diao2021similarity} built an architecture that deeply extends the attention mechanisms of image-image, text-text, and image-text tasks. For extracting high-level semantics, Li~et~al.~\cite{li2022image} utilised GCN \cite{zhang2019graph} to explore the relations of image objects. For aggregating the image and description embeddings, Chen et al. \cite{chen2021learning} proposed a special pooling operator. Wang~et~al.~\cite{wang2020stacked} proposed an end-to-end VSE network without relying on a pre-trained CNN for image feature extraction tasks.

\textbf{Methods for Finding the Underlying Meaning of Descriptions.} BERT \cite{devlin2019bert} is a supervised and widely used deep neural network for NLP tasks \cite{maltoudoglou2022well}. Singular Value Decomposition (SVD) \cite{lipovetsky2009pca} is an unsupervised matrix decomposition method and an established approach in NLP and information retrieval \cite{furnas2017information}. BERT and SVD both have dimensionality reduction capabilities that enable them to find the underlying semantic similarity between texts (e.g. sentences, image captions, documents).

\section{Proposed Semantically-Enhanced Hard Negatives Loss Function}\label{Proposed Semantically-Enhanced Hard Negatives Loss Function} 

Let $X = \begin{Bmatrix}(I_{i}, D_{i})|i=1\dots n\end{Bmatrix}$ denote a training set containing paired images and descriptions, where each image $I_{i}$ corresponds to its relevant description $D_{i}$; $i$ is the index and $n$ is the size of set $X$. Let $X^{'}=\begin{Bmatrix}(v_{i}, u_{i})|i=1\dots n\end{Bmatrix}$ be a set of image--description embedding pairs output by a VSE network, where each $i$th relevant pair consists of an image embedding $v_{i}$ and its relevant description embedding $u_{i}$. Let $\hat{v_{i}}=\begin{Bmatrix}v_{j}|j=1\dots n, j\neq i\end{Bmatrix}$ denote a set of all image embeddings from $X^{'}$ irrelevant to $u_{i}$, and $\hat{u_{i}}=\begin{Bmatrix}u_{j}|j=1\dots n, j\neq i\end{Bmatrix}$ denote a set of all description embeddings from $X^{'}$ that are irrelevant to $v_{i}$. LSH, LMH, and the proposed approach and loss function that are used during the training of VSE networks are computed as follows. 

\subsection{Related Methods and Notation}\label{Existing Loss Functions} 

\textbf{LSH Description.} The basic negatives loss function, LSH, is shown in Eq. (\ref{equationLSH}): 
\begin{eqnarray} 
\small 
\begin{aligned} 
 \mathrm{LSH}(v_{i},u_{i})=\sum_{\hat{u_{i}}}^{}[\alpha+s(v_{i},\hat{u_{i}})-s(v_{i},u_{i})]_{+} 
 +\sum_{\hat{v_{i}}}^{}[\alpha+s(u_{i},\hat{v_{i}})-s(v_{i},u_{i})]_{+} 
 \label{equationLSH} 
\end{aligned} 
\end{eqnarray} 
where $[x]_{+}\equiv \textrm{max}(x,0)$, and $\alpha$ serves as a margin parameter. Let $s(v_{i},u_{i})$ be the similarity score between the relevant image embedding $v_{i}$ and description embedding $u_{i}$; let $s(v_{i},\hat{u_{i}})$ be the set of similarity scores of the image embedding $v_{i}$ with its all irrelevant description embeddings $\hat{u_{i}}$; and let $s(u_{i},\hat{v_{i}})$ be the set of similarity scores of the description embedding $u_{i}$ with its all irrelevant image embeddings $\hat{v_{i}}$. Given a relevant pair of image--description embeddings $(v{i},u_{i})$, the result of the function takes the sum from irrelevant pairs $s(v_{i},\hat{u_{i}})$ and $s(u_{i},\hat{v_{i}})$ respectively.

\textbf{LMH Description.} The hard negatives loss function, LMH, is an improved version of LSH, that only focuses on the hard negatives \cite{faghri2018vse++}.  

\begin{eqnarray}  
\begin{aligned} 
 \mathrm{LMH}(v_{i},u_{i})=\underset{\hat{u_{i}}}{\textrm{max}}[\alpha+s(v_{i},\hat{u_{i}})-s(v_{i},u_{i})]_{+} 
 +\underset{\hat{v_{i}}}{\textrm{max}}[\alpha+s(u_{i},\hat{v_{i}})-s(v_{i},u_{i})]_{+} 
 \label{equationLMH} 
\end{aligned} 
\end{eqnarray} 
As shown in Eq. (\ref{equationLMH}), given a relevant image--description pair $(v_{i},u_{i})$, the result of the function only takes the max value of the irrelevant pairs $s(v_{i},\hat{u_{i}})$ and $s(u_{i},\hat{v_{i}})$ respectively. 

\subsection{Proposed LSEH Loss Function} \label{Proposed LSEH Loss Function} 
The proposed Semantically-Enhanced Hard negatives Loss function (LSEH) is an improved version of LMH and it is defined in Eq. (\ref{equationLSEH1}):
\begin{eqnarray}  
\begin{aligned} 
 \mathrm{LSEH}(v_{i},u_{i})= 
 \underset{\hat{u_{i}}}{\textrm{max}}[\alpha+(s(v_{i},\hat{u_{i}})+f(v_{i},\hat{u_{i}}))-s(v_{i},u_{i})]_{+} 
 \\+\underset{\hat{v_{i}}}{\textrm{max}}[\alpha+(s(u_{i},\hat{v_{i}})+f(u_{i},\hat{v_{i}}))-s(v_{i},u_{i})]_{+} 
 \label{equationLSEH1} 
\end{aligned} 
\end{eqnarray} 
LSEH introduces two sets of semantic factors $f(v_{i},\hat{u_{i}})$ and $f(u_{i},\hat{v_{i}})$ for the image--description embedding pairs $(v_{i},\hat{u_{i}})$ and the description--image embedding pairs $(u_{i},\hat{v_{i}})$ respectively, and $f(v_{i},\hat{u_{i}})$ and $f(u_{i},\hat{v_{i}})$ can be obtained via Eq. (\ref{equationfS}): 
\begin{eqnarray} 
\begin{aligned} 
    f(v_{i},\hat{u_{i}})=\lambda \times S(v_{i},\hat{u_{i}}),\; 
    f(u_{i},\hat{v_{i}})=\lambda \times S(u_{i},\hat{v_{i}}) 
    \label{equationfS} 
\end{aligned} 
\end{eqnarray} 

where $\lambda$ serves as a temperature hyperparameter, and let $S(v_{i},\hat{u_{i}})$ denote the optimal semantic similarity scores of the irrelevant image--description embedding pairs $(v_{i},\hat{u_{i}})$, and $S(u_{i},\hat{v_{i}})$ denote the optimal semantic similarity scores of the irrelevant description--image embedding pairs $(u_{i},\hat{v_{i}})$. Therefore, the question is to compute the semantic factors $f(v_{i},\hat{u_{i}})$ and $f(u_{i},\hat{v_{i}})$. 

The semantic factors are computed by finding the underlying meaning of descriptions using SVD. After pre-processing \cite{haddi2013role}, the descriptions set $\begin{Bmatrix}D_{i}|i=1\dots n\end{Bmatrix}$ is converted to a matrix $\mathrm{\mathbf{A}}$ of size $n\times w$, where $n$ is the number of descriptions, $w$ is the total number of unique terms found in the set of descriptions, and each $i$th row of $\mathrm{\mathbf{A}}$ corresponds to each $i$th description $D_{i}$. Then the truncated SVD is applied as shown in Eq. (\ref{equationSVD}) \cite{cosma2019feature}: 
\begin{eqnarray} 
\begin{aligned} 
    \mathrm{\mathbf{A}}_{n\times w} \approx \mathrm{\mathbf{U}}_{n\times k}\mathrm{\mathbf{\Lambda}}_{k\times k}\mathrm{\mathbf{V^{T}}}_{k\times w},\; 
    \mathrm{\mathbf{B}}_{n\times k} = \mathrm{\mathbf{A}}_{n\times w}\mathrm{\mathbf{V}}_{w\times k} 
    \label{equationSVD} 
\end{aligned} 
\end{eqnarray} 
where $k$ is the number of singular values. The reduced matrix $\mathrm{\mathbf{B}}$ containing $n$ rows of description vectors ($k$ dimension) is obtained by multiplying the original descriptions matrix $\mathrm{\mathbf{A}}_{n\times w}$ with matrix $\mathrm{\mathbf{V}}_{n\times k}$.  

\begin{figure}[htbp]
\centering
	\includegraphics[width=5cm]{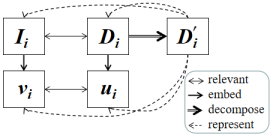}
	\caption{Based on the joint relations between $D_{i}$ with $I_{i}$, $v_{i}$, and $u_{i}$, $D^{'}_{i}$ can simultaneously represent the semantics of $I_{i}$, $D_{i}$, $v_{i}$, and $u_{i}$.}
	\label{VectorsRelations}
\end{figure}

Let a set $C=\begin{Bmatrix}D^{'}_{i}|i=1\dots n\end{Bmatrix}$ contain the reduced description vectors, and be derived from matrix $\mathrm{\mathbf{B}}_{n\times k}$, where each $i$th element $D^{'}_{i}$ is each $i$th row vector of $\mathrm{\mathbf{B}}$, therefore $D^{'}_{i}$ represents the extracted semantic of each $i$th description $D_{i}$. Also, as shown in Fig. \ref{VectorsRelations}, description $D_{i}$ is relevant to image $I_{i}$, and embeddings $u_{i}$ and $v_{i}$ are output from $D_{i}$ and $I_{i}$ respectively, hence $D^{'}_{i}$ can also simultaneously represent the semantics of $I_{i}$, $v_{i}$, and $u_{i}$. 

Therefore, let $\hat{D^{'}_{i}}=\begin{Bmatrix}D^{'}_{j}|j=1\dots n, j\neq i\end{Bmatrix}$ denote a set of reduced vectors of descriptions from set $C$, where each $j$th vector $D^{'}_{j}$ simultaneously represents the optimal semantics of $v_{j}$ and $u_{j}$ from sets $\hat{v_{i}}$ and $\hat{u_{i}}$ respectively, then $S(v_{i},\hat{u_{i}})$ and $S(u_{i},\hat{v_{i}})$ can be alternatively calculated using Eq. (\ref{equationSD}): 
\begin{eqnarray} 
    S(v_{i},\hat{u_{i}})=S(u_{i},\hat{v_{i}})=s(D^{'}_{i}, \hat{D^{'}_{i}}) 
    \label{equationSD} 
\end{eqnarray} 
where $s(D^{'}_{i}, \hat{D^{'}_{i}})\in[-1,1]$ computes a set of cosine similarity scores of $D^{'}_{i}$ with $\hat{D^{'}_{i}}$, thus: 
\begin{eqnarray} 
    f(v_{i},\hat{u_{i}})=f(u_{i},\hat{v_{i}})=\lambda \times s(D^{'}_{i}, \hat{D^{'}_{i}}) 
    \label{equationB} 
\end{eqnarray} 
Furthermore, as a recent popular pre-trained language model, BERT \cite{devlin2019bert} can also find the underlying meaning of descriptions. BERT relies on a large amount of data for training and is known to perform well in processing long documents \cite{devlin2019bert}. The experiment in Section \ref{Comparing SVD and BERT} compares the performance of LSEH when using SVD and when using BERT. 
Finally, integrating Eq. (\ref{equationB}) into Eq. (\ref{equationLSEH1}), LSEH is computed as Eq. (\ref{equationLSEH}): 
\begin{eqnarray} 
\begin{aligned} 
 \mathrm{LSEH}(v_{i},u_{i})= 
 \underset{\hat{u_{i}}}{\textrm{max}}[\alpha+s(v_{i},\hat{u_{i}})+\lambda \times s(D^{'}_{i}, \hat{D^{'}_{i}})-s(v_{i},u_{i})]_{+} 
 \\+\underset{\hat{v_{i}}}{\textrm{max}}[\alpha+s(u_{i},\hat{v_{i}})+\lambda \times s(D^{'}_{i}, \hat{D^{'}_{i}})-s(v_{i},u_{i})]_{+} 
 \label{equationLSEH} 
\end{aligned} 
\end{eqnarray} 
 
\begin{figure}[htbp]
\centering
	\includegraphics[width=10cm]{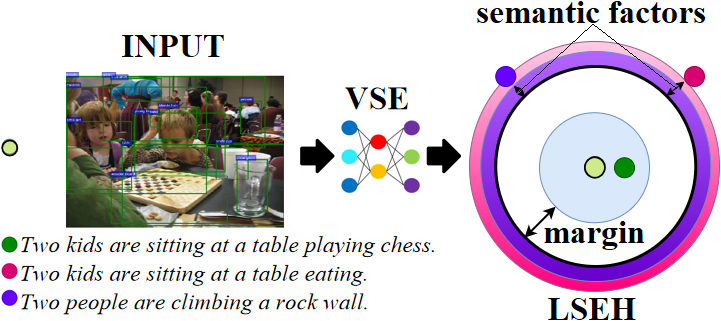}
	\caption{Illustrations of LSEH. LSEH dynamically adjusts the learning objective for the VSE network, and taking the maximum gradients from the ambiguous image--description pairs.}
	\label{tripletloss}
\end{figure}

\textbf{Illustration of LSEH.} 
As shown in Fig. \ref{tripletloss}, the learning objective of LSEH, defined by margin $\alpha$, is dynamically adjusted for every irrelevant image--description pair based on their optimal semantic similarity (see Eq. (\ref{equationB})).  
LSEH has two purposes: (1) it dynamically adjusts the learning objective for the VSE network for flexible learning; (2) performs efficient training by taking the maximum gradients from the ambiguous image--description pairs because their large optimal similarity scores that cause the large gradients. 

\subsection{Process of Training VSE Networks using LSEH} 
Algorithm (\ref{LSEHtrainsVSE}) shows the process of using LSEH to train a VSE network. Initially, the descriptions from the train set $X$ are represented as a set of reduced vectors $C$ (line 1). Thereafter the VSE network is applied as follows (line 2). In every epoch, the train set $X$ is split into mini-batches and then the corresponding set of vectors are obtained from $C$ such that $\begin{Bmatrix}(X, C)=(X_{p}, C_{p})|p=1\dots m\end{Bmatrix}$, where $p$ is the index and $m$ is the number of mini-batches (lines 3-4). 

For each mini-batch $(X_{p}, C_{p})$, the network outputs a set $X^{'}_{p}$ of embeddings from $X_{p}$ (lines 5-6). Thereafter, for every relevant image--description embedding pair $(v_{i},u_{i})$ in set $X^{'}_{p}$, the semantic factors are computed using set $C_{p}$ (Eq. (\ref{equationB})), then the LSEH value is computed (Eq. (\ref{equationLSEH})) (lines 7-10). Finally, the LSEH value from the mini-batch is used for backpropagation (line 11). Furthermore, if the number of mini-batches reaches the validation step $Vstep$, the network is validated and the model that achieved the best overall performance is saved (lines 12-17). 
\begin{algorithm}[htbp]
\small
\caption{Pseudocode of training a VSE network using LSEH}
\label{LSEHtrainsVSE}
\hspace*{0.01in} {\bf Input:} 
 $X$ training set (image--description pairs),\\
\hspace*{0.39in} {$E$ training epochs, $Vstep$ validation step}\\
\hspace*{0.01in} {\bf Output:}
LSEH
    \begin{algorithmic}[1]
	  \STATE Obtain set $C$ based on set $X$ $\triangleright$ Formula \ref{equationSVD} \STATE Apply the VSE network
	  \FOR{\uline{epoch in $E$}}
	     \STATE Split sets $(X,C)$ to mini-batches
		 \FOR{\uline{mini-batch $(X_{p},C_{p})$ in $(X,C)$}}
		     \STATE Embed set $X_{p}$ to set $X^{'}_{p}$
		     \FOR{\uline{$v_{i},u_{i}$ in $X^{'}_{p}$}}
		          \STATE Use $C_{p}$ to compute the semantic factors $\triangleright$ Formula \ref{equationB}
		          \STATE Use the semantic factors to compute LSEH $\triangleright$ Formula \ref{equationLSEH}
		     \ENDFOR
		     \STATE Backpropagate LSEH for updating network
		     \IF{\uline{mini-batches number == $Vstep$}}
		          \STATE Validate the model
		          \IF{\uline{the overall performance is the best}}
		              \STATE Save the current model
		          \ENDIF
		     \ENDIF
		 \ENDFOR
	  \ENDFOR
	\end{algorithmic}
\end{algorithm}

\section{Experiments} 
\subsection{Experiment Setup} 

\textbf{Datasets and Protocols.} The datasets utilised in the experiments are Flickr30K~\cite{young2014image} and MS-COCO \cite{lin2014microsoft}, and these datasets are typically used for evaluating the performance of VSE networks \cite{faghri2018vse++}. This paper also utilises the IAPR TC-12 dataset \cite{grubinger2006iapr}. The datasets were split into train, test and validation sets as shown in Table \ref{DatasetsSplit} \cite{faghri2018vse++}. In the Flickr30K and MS-COCO datasets, every image is associated with five relevant textual descriptions. IAPR TC-12 includes pictures of different sports and actions, photographs of people, animals, cities, landscapes and many other aspects of contemporary life. Each image in IAPR TC-12 is associated with one relevant textual description \cite{grubinger2006iapr}. 

\begin{table}[htbp]
\centering
\caption{Dataset split of Flickr30K, MS-COCO, and IAPR TC-12.}
\label{DatasetsSplit}
\begin{tabular}{cccc}
\hline
Dataset & Train & Validate & Test \\ \hline
Flickr30K & 29000 & 1014 & 1000 \\
MS-COCO & 113287 & 1000 & 1000 and 5000 \\
IAPRTC-12 & 18000 & 1000 & 1000 \\ \hline
\end{tabular}
\end{table}

The descriptions were pre-processed with lowercase, stemming, removing punctuation and alphabetic, filtering out stop words and short words, and the term frequency-inverse document frequency (TFIDF) text vectorizer was applied to transform the descriptions into usable vectors \cite{martineau2009delta}. The number of singular values $k$ for SVD was set as 400 for all datasets. The process of representing descriptions in a reduced dimensional space is described in Section \ref{Proposed Semantically-Enhanced Hard Negatives Loss Function}. 
SVD is computed once for each dataset. LSEH takes the vector from matrix $\mathrm{\mathbf{B}}_{n\times k}$ (see Eq. (\ref{equationSVD})) to compute the cosine similarity between vectors from the matrix. Therefore, the cost for LSEH is only the cosine similarity computation. Section \ref{Comparisons on Computation Time} compares LSEH, LMH, and LUWM on computation time.  

\textbf{Network Implementations.}\label{Experiments Methodology} 
The experiments tested VSE++\cite{ faghri2018vse++}, VSRN \cite{li2022image}, SGRAF (sub-networks SGRAF-SAF and SGRAF-SGR) \cite{diao2021similarity}, VSE$\infty$\cite{chen2021learning}, and GSMN \cite{liu2020graph}. For consistency of comparisons, VSE++, VSRN, SGRAF, VSE$\infty$, and GSMN have been tuned using the image region feature (size of 36$\times$2048) that was pre-extracted by the modified faster R-CNN \cite{anderson2018bottom}, where the architecture of the faster R-CNN includes a backbone of ResNet-101 that was pre-trained on ImageNet \cite{krizhevsky2012imagenet} and Visual Genome \cite{krishna2017visual}.  
The architectures of VSE++, VSRN, and SGRAF (sub-networks SGRAF-SAF and SGRAF-SGR) follow those described in \cite{faghri2018vse++}, \cite{li2022image}, \cite{diao2021similarity} respectively. 
For VSE$\infty$, the image region mode of architecture was used, and it follows the one described in \cite{chen2021learning}.  
For GSMN the sparse mode of architecture was used, and it follows the one described in \cite{liu2020graph}. 
The source-code files of the abovementioned networks when using the proposed LSEH loss function and their loss functions are provided in our GitHub repository \footnote{https://github.com/yangong23/VSEnetworksLSEH}. 

\textbf{Network Training Hyperparameters.} The benchmark hyperparameter settings for each network are described in \cite{faghri2018vse++, wei2021universal, li2022image, diao2021similarity, chen2021learning} respectively, where the benchmark hyperparameter settings on the new IAPRTC-12 refer to Flickr30K because they contain a similar number of images. Let LMH1 denote LMH with the benchmark hyperparameters of each network and LMH2 denote LMH with LSEH's hyperparameter settings, then all the hyperparameters except those shown in Table \ref{Hyperparameters} for using LMH1, LMH2, LUWM, and LSEH refer to each network's benchmark-settings.  

The hyperparameters shown in Table \ref{Hyperparameters} were selected experimentally and tuned as follows: (1) for LSEH the learning and update learning rates can be set to a larger value at an earlier epoch than for LMH1; (2) LSEH sets the margin $\alpha$ as 0.185 and the semantic factor temperature hyperparameter $\lambda$ as 0.025. All of the experiments were conducted on a workstation with NVIDIA Titan GPU. For a fair comparison, all software programs were set with the same random seed. 

\begin{table*}[htbp]
\scriptsize
\caption{Hyperparameters with different settings for LMH1, LMH2, LSEH, and LUWM, in terms of Margin, Learning Rate, and epochs to update learning rate (Update).}
\label{Hyperparameters}
\centering 
\resizebox{\textwidth}{!}{
\begin{tabular}{lccclccclccc}
\hline
              & \multicolumn{3}{c}{Flickr30K}   &  & \multicolumn{3}{c}{MS-COCO}     &  & \multicolumn{3}{c}{IAPR TC-12}  \\ \cline{2-4} \cline{6-8} \cline{10-12} 
Loss & Margin & LR & Update &  & Margin & LR & Update &  & Margin & LR & Update \\ \hline
\multicolumn{12}{c}{VSE++}                                                                                                \\ \hline
LMH1          & 0.20   & 0.0002        & 15     &  & 0.20   & 0.0002        & 15     &  & 0.20   & 0.0002        & 25     \\
LMH2          & 0.16   & 0.0020        & 3      &  & 0.16   & 0.0030        & 4      &  & 0.16   & 0.0008        & 5      \\
LSEH          & 0.16-0.21   & 0.0020        & 3      &  & 0.16-0.21   & 0.0030        & 4      &  & 0.16-0.21   & 0.0008        & 5      \\ \hline
\multicolumn{12}{c}{VSRN}                                                                                                 \\ \hline
LMH1          & 0.20   & 0.0002        & 10     &  & 0.20   & 0.0002        & 15     &  & 0.20   & 0.0005        & 20     \\
LMH2          & 0.16   & 0.0004        & 5      &  & 0.16   & 0.0004        & 6      &  & 0.16   & 0.0005        & 20     \\
LSEH          & 0.16-0.21   & 0.0004        & 5      &  & 0.16-0.21   & 0.0004        & 6      &  & 0.16-0.21   & 0.0005        & 20     \\ \hline
\multicolumn{12}{c}{SGRAF-SAF}                                                                                            \\ \hline
LMH1          & 0.20   & 0.0002        & 20     &  & 0.20   & 0.0002        & 10     &  & 0.20   & 0.0002        & 20     \\
LMH2          & 0.16   & 0.0004        & 16     &  & 0.16   & 0.0004        & 6      &  & 0.16   & 0.0004        & 15     \\
LSEH          & 0.16-0.21   & 0.0004        & 16     &  & 0.16-0.21   & 0.0004        & 6      &  & 0.16-0.21   & 0.0004        & 15     \\ \hline
\multicolumn{12}{c}{SGRAF-SGR}                                                                                            \\ \hline
LMH1          & 0.20   & 0.0002        & 30     &  & 0.20   & 0.0002        & 10     &  & 0.20   & 0.0002        & 30     \\
LMH2          & 0.16   & 0.0004        & 20     &  & 0.16   & 0.0004        & 6      &  & 0.16   & 0.0004        & 20     \\
LSEH          & 0.16-0.21   & 0.0004        & 20     &  & 0.16-0.21   & 0.0004        & 6      &  & 0.16-0.21   & 0.0004        & 20     \\ \hline
\multicolumn{12}{c}{VSE$\infty$}                                                                                          \\ \hline
LMH1          & 0.20   & 0.0005        & 15     &  & 0.20   & 0.0005        & 15     &  & 0.20   & 0.0005        & 15     \\
LMH2          & 0.16   & 0.0008        & 10     &  & 0.16   & 0.0010        & 10     &  & 0.16   & 0.0009        & 10     \\
LSEH          & 0.16-0.21   & 0.0008        & 10     &  & 0.16-0.21   & 0.0010        & 10     &  & 0.16-0.21   & 0.0009        & 10     \\ \hline
\multicolumn{12}{c}{GSMN}                                                                                                 \\ \hline
LMH1          & 0.20   & 0.0002        & 15     &  & 0.20   & 0.0005        & 5      &  & $-$    & $-$           & $-$    \\
LUWM          & $-$    & 0.0002        & 15     &  & $-$    & 0.0005        & 5      &  & $-$    & $-$           & $-$    \\
LSEH          & 0.16-0.21   & 0.0004        & 10     &  & 0.16-0.21   & 0.0006        & 4      &  & $-$    & $-$           & $-$    \\ \hline
\end{tabular}
 }
\end{table*}

\textbf{Evaluation Measures.} To evaluate the performance of the networks two evaluation measures were used: Recall@k and M-Recall.\\ 
\textbf{Recall@k}. The evaluation measure for the cross-modal information retrieval experiments  is the commonly used Recall at rank $k$ (Recall@$k$), which is defined as the percentage of relevant items in the top $k$ retrieved results \cite{faghri2018vse++}. 
The experiments evaluate the performance of the network in retrieving any one of the relevant items from the list of relevant items and computed the average Recall of the results of the test queries \cite{faghri2018vse++}.\\ 
\textbf{M-Recall}. Defined as Eq. (\ref{MeanRecall}) \cite{faghri2018vse++}: 
\begin{eqnarray} 
\begin{aligned} 
 \mathrm{M\raisebox{0mm}{-}Recall}=\frac{1}{6}\sum_{k}^{1,5,10}(\mathrm{Recall}_{I2T}@k+\mathrm{Recall}_{T2I}@k) 
 \label{MeanRecall} 
\end{aligned} 
\end{eqnarray} 
where M-Recall is the Mean of average Recall@1, 5, and 10 from both image-to-text ($I2T$) and text-to-image ($T2I$) retrieval. M-Recall is used for evaluating the overall performance of a network during the validation stages. 

\subsection{Comparison of LMH and LSEH on Training Efficiency} \label{ResultsEfficiency} 
This section compares the learning performance of VSE++, VSRN, SGRAF-SAF, SGRAF-SGR, and VSE$\infty$ when using LMH1, LMH2, and LSEH using the train and validation sets of the Flickr30K, MS-COCO, and IAPR TC-12 datasets. 
Obtaining the optimal trained model for the VSE networks is through validating the middle-trained model on the overall performance using the M-Recall evaluation measure. Validation occurs at every 1000 mini-batches for SGRAF \cite{diao2021similarity}, and at every 500 mini-batches for the VSE$\infty$ \cite{chen2021learning}, VSRN \cite{li2022image}, and VSE++ \cite{faghri2018vse++}. 

\textbf{Graphic and Quantitative Results.}
\begin{figure*}[htbp]
\centering
	\includegraphics[width=0.9\textwidth]{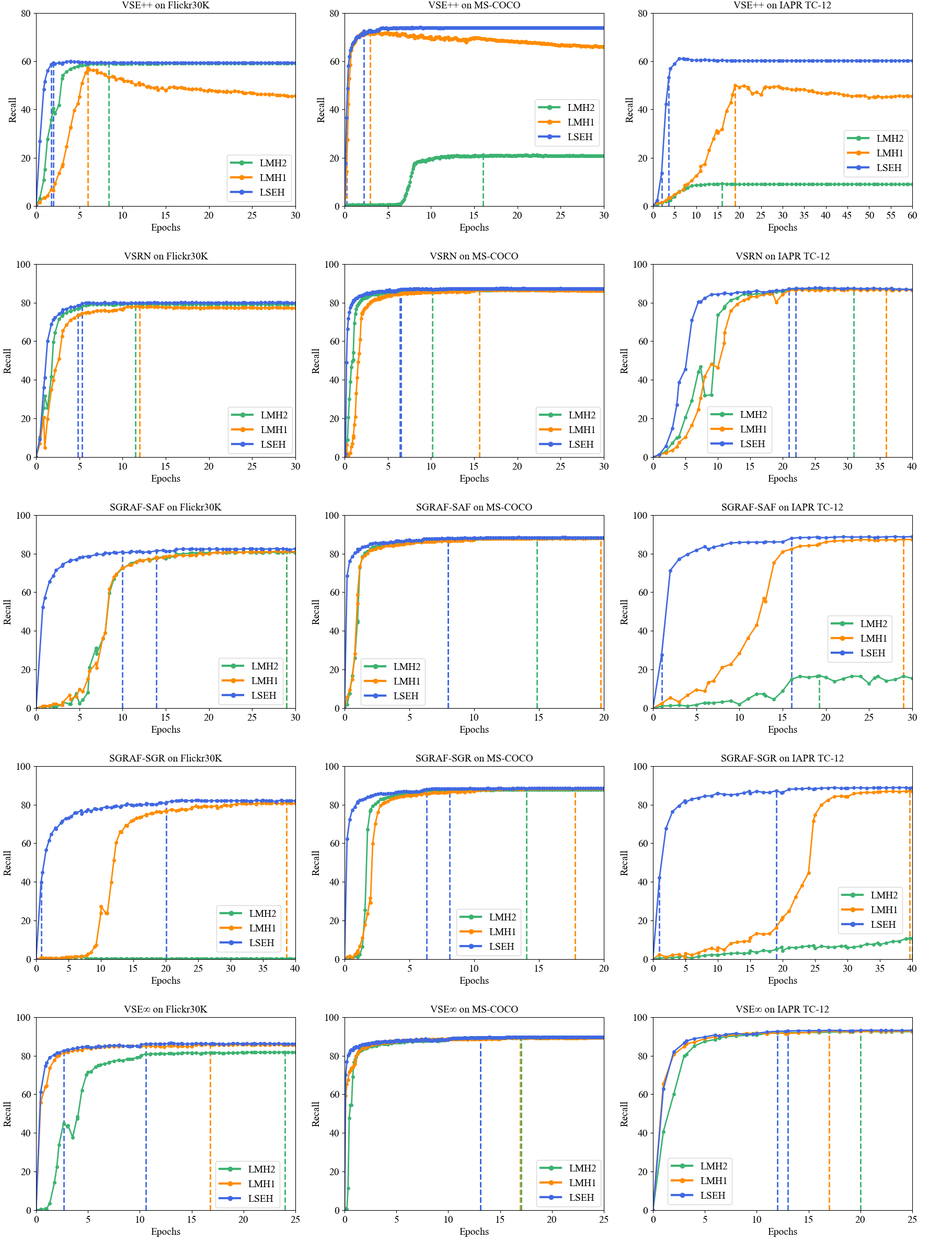}
	\caption{\small Comparison of training efficiency when using LMH1, LMH2, and LSEH to train various VSE networks on Flickr30K, MS-COCO, and IAPR TC-12. Recall (in Y-axis) is M-Recall (\%). The largest M-Recall values of LMH1 and LMH2 are denoted by the vertical dashed lines of orange and green respectively. The vertical dashed blue lines indicate that LSEH's M-Recall has exceeded the best performance of LMH1 or LHM2.}
	\label{figEfficiency}
\end{figure*}
Fig. \ref{figEfficiency} compares the training performance of networks using LMH1, LMH2, and LSEH. The comparison considers the number of epochs that each network needs to reach its largest M-Recall value. The comparison is illustrated in Fig. \ref{figEfficiency}. 
Table \ref{tableEfficiency} quantifies the improvement in training efficiency of each network when using LMH and the proposed LSEH loss functions. In Table \ref{tableEfficiency}, for the comparison of LMH1 and LSEH $\textrm{M-Recall}_{\textrm{LMH1}}$ is LMH1's largest M-Recall value. $\textrm{epochs}_{\textrm{LMH1}}$ is the number of epochs needed to achieve $\textrm{M-Recall}_{\textrm{LMH1}}$ by LMH1. $\textrm{epochs}_{\textrm{LSEH}}$ is the number of epochs needed to achieve $\textrm{M-Recall}_{\textrm{LMH1}}$ by LSEH. Same interpretation applies for the comparison of LMH2 and LSEH (right side of Table \ref{tableEfficiency}). 

Difference is computed using Eq. (\ref{Diff}). 
\begin{eqnarray} 
\begin{aligned} 
 \textrm{Difference}=(\frac{\textrm{epochs}_{\textrm{LSEH}}-\textrm{epochs}_{\textrm{LMH}}}{\textrm{epochs}_{\textrm{LMH}}})\%, 
 \label{Diff} 
\end{aligned} 
\end{eqnarray} where LMH is either LMH1 or LMH2. A negative Difference value denotes an improvement in performance when using LSEH. 
As shown in Table \ref{tableEfficiency}, when using LSEH, the epochs needed to exceed the largest M-Recall values with LMH1 and LMH2 could be reduced by 53.2\% and 74.7\% on Flickr30K, by 43.0\% and 52.1\% on MS-COCO, and by 48.5\% and 69.8\% on IAPR TC-12 respectively with LSEH. The largest improvement was for VSE++, where the training efficiency on IAPR TC-12 was improved by approximately 80.5\% with LSEH. 

\begin{table*}[!htbp] 
 \caption{Comparison of training efficiency of each network when using LMH and the proposed LSEH loss function. Difference is calculated as shown in (\ref{Diff}).}
 \label{tableEfficiency}
 \centering 
\resizebox{\linewidth}{!}{
\begin{tabular}{lcccrlcccr}
\hline
            & \multicolumn{4}{c}{Comparison of LMH1 and LSEH}                                                                                              &  & \multicolumn{4}{c}{Comparison of LMH2 and LSEH}                                                                                              \\ \cline{2-5} \cline{7-10} 
Network     & \begin{tabular}[c]{@{}c@{}}M-Recall\\ LMH1\end{tabular} & \begin{tabular}[c]{@{}c@{}}epochs\\ LMH1\end{tabular} & \begin{tabular}[c]{@{}c@{}}epochs\\ LSEH\end{tabular} & \multicolumn{1}{c}{Difference} &  & \begin{tabular}[c]{@{}c@{}}M-Recall\\ LMH2\end{tabular} & \begin{tabular}[c]{@{}c@{}}epochs\\ LMH2\end{tabular} & \begin{tabular}[c]{@{}c@{}}epochs\\ LSEH\end{tabular} & \multicolumn{1}{c}{Difference} \\ \hline
\multicolumn{10}{c}{Flickr30K}                                                                                                                                                                                                                                                                               \\ \hline
VSE++       & 57.1                                & 6.0                               & 1.8                               & -4.2 (-70.0\%)                 &  & 59.1                                & 8.4                               & 2.0                               & -6.4 (-76.2\%)                 \\
VSRN        & 78.0                                & 12.0                              & 4.9                               & -7.1 (-59.2\%)                 &  & 79.4                                & 11.5                              & 5.3                               & -6.2 (-53.9\%)                 \\
SGRAF-SAF   & 81.4                                & 29.0                              & 13.9                              & -15.1 (-52.1\%)                &  & 80.9                                & 29.0                              & 10.0                              & -19.0 (-65.5\%)                \\
SGRAF-SGR   & 81.0                                & 38.6                              & 20.1                              & -18.5 (-47.9\%)                &  & 1.6                                 & 7.0                               & 0.8                               & -6.2 (-88.6\%)                 \\
VSE$\infty$ & 85.9                                & 16.8                              & 10.6                              & -6.2 (-36.9\%)                 &  & 81.8                                & 24.0                              & 2.6                               & -21.4 (-89.2\%)                \\ \hline
Average     &                                     &                                   &                                   & -10.2 (-53.2\%)                &  &                                     &                                   &                                   & -11.8 (-74.7\%)                \\ \hline
\multicolumn{10}{c}{MS-COCO}                                                                                                                                                                                                                                                                                 \\ \hline
VSE++       & 72.5                                & 2.9                               & 2.3                               & -0.6 (-20.7\%)                 &  & 21.3                                & 16.0                              & 0.2                               & -15.8 (-98.8\%)                \\
VSRN        & 86.8                                & 15.6                              & 6.6                               & -9.0 (-57.7\%)                 &  & 86.7                                & 10.2                              & 6.4                               & -3.8 (-37.3\%)                 \\
SGRAF-SAF   & 88.0                                & 19.8                              & 8.0                               & -11.8 (-59.6\%)                &  & 88.0                                & 14.8                              & 8.0                               & -6.8 (-45.9\%)                 \\
SGRAF-SGR   & 88.4                                & 17.8                              & 8.1                               & -9.7 (-54.5\%)                 &  & 87.8                                & 14.0                              & 6.3                               & -7.7 (-55.0\%)                 \\
VSE$\infty$ & 89.4                                & 16.9                              & 13.1                              & -3.8 (-22.5\%)                 &  & 89.5                                & 17.1                              & 13.1                              & -4.0 (-23.4\%)                 \\ \hline
Average     &                                     &                                   &                                   & -7.0 (-43.0)\%                 &  &                                     &                                   &                                   & -7.6 (-52.1\%)                 \\ \hline
\multicolumn{10}{c}{IAPR TC-12}                                                                                                                                                                                                                                                                              \\ \hline
VSE++       & 50.0                                & 19.0                              & 3.7                               & -15.3 (-80.5\%)                &  & 9.3                                 & 16.0                              & 2.0                               & -14.0 (-87.5\%)                \\
VSRN        & 86.7                                & 36.0                              & 21.0                              & -15.0 (-41.7\%)                &  & 87.2                                & 31.0                              & 22.0                              & -9.0 (-29.0\%)                 \\
SGRAF-SAF   & 87.3                                & 29.0                              & 16.0                              & -13.0 (-44.8\%)                &  & 16.8                                & 19.2                              & 1.0                               & -18.2 (-94.8\%)                \\
SGRAF-SGR   & 87.1                                & 39.6                              & 19.0                              & -20.6 (-52.0\%)                &  & 10.9                                & 40.0                              & 1.0                               & -39.0 (-97.5\%)                \\
VSE$\infty$ & 92.7                                & 17.0                              & 13.0                              & -4.0 (-23.5\%)                 &  & 92.5                                & 20.0                              & 12.0                              & -8.0 (-40.0\%)                 \\ \hline
Average     &                                     &                                   &                                   & -13.6 (-48.5\%)                &  &                                     &                                   &                                   & -17.6 (-69.8\%)                \\ \hline
\end{tabular}
}
\end{table*}

\subsection{Comparison of LMH and LSEH on Cross-modal Information Retrieval} 
This section compares the cross-modal information retrieval performance between the optimal models of VSE++, VSRN, SGRAF-SAF, SGRAF-SGR, and VSE$\infty$ with using LMH1, LMH2, and LSEH on the Flickr30K, MS-COCO 1K, MS-COCO 5K, and IAPR TC-12 test sets. Table \ref{tablePerformance1}, \ref{tablePerformance2}, \ref{tablePerformance3}, and \ref{tablePerformance4} show the results of the comparisons of LMH1, LMH2, and LSEH when integrated into various networks and applied to the datasets of Flickr30K (see Table \ref{tablePerformance1}), MS-COCO 1K (see Table \ref{tablePerformance2}), MS-COCO 5K (see Table \ref{tablePerformance3}), and IAPR TC-12 (see Table \ref{tablePerformance4}) respectively. Recall values @1, @5, and @10 show the average Recall values across all queries found in each of the test sets (see Table \ref{DatasetsSplit}). The last column of Tables \ref{tablePerformance1}, \ref{tablePerformance2}, \ref{tablePerformance3}, and \ref{tablePerformance4} shows the mean of the average Recall values (i.e. the average of columns 3-5).  Table \ref{tablePerformanceAve} summarises the mean average Recall across five networks for each dataset from Table \ref{tablePerformance1}, \ref{tablePerformance2}, \ref{tablePerformance3}, and \ref{tablePerformance4}, and below is a summary of the main findings for each dataset. \\\textbf{(1)~Flickr30K}. The proposed LSEH reached a Recall of 85.2\% for image-to-text, and a Recall of 73.0\% for text-to-image retrieval. LSEH outperformed LMH1 by 2.3\% and 2.0\% for image-to-text and text-to-image retrieval, respectively. LSEH outperformed LMH2 by 19.6\% for image-to-text retrieval and by 16.7\% for text-to-image retrieval. \\\textbf{(2)~MS-COCO1K}. For LSEH, Recall reached 89.0\% for image-to-text retrieval and 80.5\% for text-to-image retrieval, and outperformed LMH1 by 0.2\% and 0.2\% for those tasks, respectively. LSEH also outperformed LMH2 for image-to-text and text-to-image retrieval by 9.3\% and 12.2\%, respectively. \\\textbf{(3)~MS-COCO5K}. For image-to-text retrieval, LSEH's Recall reached 71.7\% which outperformed LMH1 by 0.7\% and LMH2 by 8.5\%. For text-to-image retrieval, the Recall of the proposed LSEH reached 59.7\% which outperformed LMH1 by 0.3\% and LMH2 by 8.6\%. \\\textbf{(4)~IAPR TC-12}. LSEH's Recall values for image-to-text and text-to-image retrieval were 84.5\% and 84.3\% respectively, and these outperformed the results of LMH1 by 3.5\% and 2.8\% respectively. The proposed LSEH outperformed LMH2 for image-to-text and text-to-image retrieval by 38.6\% and 44.5\% respectively. 

\begin{table*}[htbp]
\caption{Results of cross-modal information retrieval by LMH1, LMH2, and LSEH on the Flickr30K dataset in terms of average Recall@$k$ (\%).}
\label{tablePerformance1}
\centering
\resizebox{\linewidth}{!}{
\begin{tabular}{lccccrlcccr}
\hline
            & \multicolumn{1}{l}{}              & \multicolumn{4}{c}{Image-to-Text Retrieval}                                                                                 &  & \multicolumn{4}{c}{Text-to-Image Retrieval}                                                                                                   \\ \cline{3-6} \cline{8-11}
Network     & \multicolumn{1}{l}{Loss} & R@1                       & R@5                       & R@10                      & \multicolumn{1}{c}{Mean} &  & R@1                       & R@5                       & R@10                      & \multicolumn{1}{c}{Mean}                   \\ \hline
VSE++       & LMH1                              & 38.9                           & 68.0                           & 78.5                           & 61.8 ($\pm$16.8)         &  & 29.7                           & 58.7                           & 69.8                           & 52.7 ($\pm$16.9)                           \\
            & LMH2                              & 41.3                           & 71.0                           & 82.2                           & 64.8 ($\pm$17.3)         &  & \textbf{33.3} & 61.5                           & 72.0                           & 55.6 ($\pm$16.3)                           \\
            & LSEH                              & \textbf{45.9} & \textbf{74.0} & \textbf{82.7} & \textbf{67.5 ($\pm$15.7)}         &  & 33.2                           & \textbf{62.2} & \textbf{73.3} & \textbf{56.2 ($\pm$16.9)} \\ \hline
VSRN        & LMH1                              & 69.8                           & 89.0                           & 94.4                           & 84.4 ($\pm$10.6)         &  & 52.1                           & 78.8                           & 86.6                           & 72.5 ($\pm$14.8)                           \\
            & LMH2                              & 71.1                           & 91.5                           & \textbf{95.8} & 86.1 ($\pm$10.8)         &  & 54.8                           & 81.0                           & 87.4                           & 74.4 ($\pm$14.1)                           \\
            & LSEH                              & \textbf{73.0} & \textbf{92.8} & 95.7                           & \textbf{87.2 ($\pm$10.1)}         &  & \textbf{55.8} & \textbf{81.9} & \textbf{88.8} & \textbf{75.5 ($\pm$14.2)} \\ \hline
SGRAF-SAF   & LMH1                              & 75.5                           & 93.6                           & 97.0                           & 88.7 ($\pm$9.4)          &  & 55.3                           & 81.7                           & 88.5                           & 75.2 ($\pm$14.3)                           \\
            & LMH2                              & 74.0                           & 93.6                           & 97.0                           & 88.2 ($\pm$10.1)         &  & 54.6                           & 80.6                           & 87.0                           & 74.1 ($\pm$14.0)                           \\
            & LSEH                              & \textbf{76.2} & \textbf{93.8} & \textbf{97.2} & \textbf{89.1 ($\pm$9.2)}          &  & \textbf{57.7} & \textbf{82.3} & \textbf{88.8} & \textbf{76.3 ($\pm$13.4)} \\ \hline
SGRAF-SGR   & LMH1                              & 74.2                           & 92.5                           & 96.5                           & 87.7 ($\pm$9.7)          &  & 55.4                           & 80.6                           & 85.9                           & 74.0 ($\pm$13.3)                           \\
            & LMH2                              & 0.3                            & 1.3                            & 2.0                            & 1.2 ($\pm$0.7)           &  & 0.4                            & 1.5                            & 2.8                            & 1.6 ($\pm$1.0)                             \\
            & LSEH                              & \textbf{78.2} & \textbf{94.3} & \textbf{96.8} & \textbf{89.8 ($\pm$8.2)}          &  & \textbf{57.6} & \textbf{82.5} & \textbf{87.9} & \textbf{76.0 ($\pm$13.2)} \\ \hline
VSE$\infty$ & LMH1                              & 80.8                           & \textbf{96.4} & 98.3                           & 91.8 ($\pm$7.8)          &  & 62.6                           & 86.9                           & 91.7                           & 80.4 ($\pm$12.7)                           \\
            & LMH2                              & 74.4                           & 91.7                           & 96.3                           & 87.5 ($\pm$9.4)          &  & 56.1                           & 82.3                           & 89.0                           & 75.8 ($\pm$14.2)                           \\
            & LSEH                              & \textbf{82.4} & 96.0                           & \textbf{98.6} & \textbf{92.3 ($\pm$7.1)}          &  & \textbf{63.7} & \textbf{87.1} & \textbf{92.5} & \textbf{81.1 ($\pm$12.5)} \\ \hline
            
\end{tabular}
}
\end{table*}

\begin{table*}[htbp]
\caption{Results of cross-modal  information retrieval by LMH1, LMH2, and LSEH on the MS-COCO1K dataset in terms of average Recall@$k$ (\%).}
\label{tablePerformance2}
\centering
\resizebox{\linewidth}{!}{
\begin{tabular}{lccccrlcccr}
\hline
            & \multicolumn{1}{l}{}              & \multicolumn{4}{c}{Image-to-Text Retrieval}                                                                                 &  & \multicolumn{4}{c}{Text-to-Image Retrieval}                                                                                                   \\ \cline{3-6} \cline{8-11}
Network     & \multicolumn{1}{l}{Loss} & R@1                       & R@5                       & R@10                      & \multicolumn{1}{c}{Mean} &  & R@1                       & R@5                       & R@10                      & \multicolumn{1}{c}{Mean}                   \\ \hline
VSE++       & LMH1                              & 54.5                           & 84.9                           & 91.7                           & 77.0 ($\pm$16.2)         &  & 40.8                           & 75.6                           & 86.3                           & 67.6 ($\pm$19.4)                           \\
            & LMH2                              & 11.9                           & 38.5                           & 53.0                           & 34.5 ($\pm$17.0)         &  & 2.5                            & 8.6                            & 13.8                           & 8.3 ($\pm$4.6)                             \\
            & LSEH                              & \textbf{56.3} & \textbf{85.3} & \textbf{92.2} & \textbf{77.9 ($\pm$15.6)}         &  & \textbf{41.2} & \textbf{76.1} & \textbf{86.6} & \textbf{68.0 ($\pm$19.4)} \\ \hline
VSRN        & LMH1                              & 76.4                           & 94.2                           & 97.6                           & 89.4 ($\pm$9.3)          &  & \textbf{63.1} & 89.4                           & 94.3                           & 82.3 ($\pm$13.7)                           \\
            & LMH2                              & 73.5                           & 94.7                           & 98.1                           & 88.8 ($\pm$10.9)         &  & 61.7                           & 89.3                           & 94.8                           & 81.9 ($\pm$14.5)                           \\
            & LSEH                              & \textbf{76.5} & \textbf{95.7} & \textbf{98.4} & \textbf{90.2 ($\pm$9.7)}          &  & 62.0                           & \textbf{90.1} & \textbf{95.1} & \textbf{82.4 ($\pm$14.6)} \\ \hline
SGRAF-SAF   & LMH1                              & \textbf{80.2} & 96.6                           & 98.7                           & 91.8 ($\pm$8.3)          &  & 64.5                           & 90.5                           & 96.0                           & 83.7 ($\pm$13.7)                           \\
            & LMH2                              & 78.8                           & 96.3                           & 98.6                           & 91.2 ($\pm$8.8)          &  & 64.0                           & 90.6                           & 95.5                           & 83.4 ($\pm$13.8)                           \\
            & LSEH                              & 80.1                           & \textbf{97.3} & \textbf{98.7} & \textbf{92.0 ($\pm$8.5)}          &  & \textbf{64.6} & \textbf{90.9} & \textbf{96.0} & \textbf{83.8 ($\pm$13.8)} \\ \hline
SGRAF-SGR   & LMH1                              & 79.7                           & 97.0                           & 98.8                           & 91.8 ($\pm$8.6)          &  & \textbf{64.0} & 90.5                           & 95.7                           & 83.4 ($\pm$13.9)                           \\
            & LMH2                              & 79.7                           & 96.3                           & 98.6                           & 91.5 ($\pm$8.4)          &  & 63.7                           & 90.5                           & 95.5                           & 83.2 ($\pm$14.0)                           \\
            & LSEH                              & \textbf{80.1} & \textbf{97.7} & \textbf{99.1} & \textbf{92.3 ($\pm$8.6)}          &  & 63.9                           & \textbf{90.7} & \textbf{95.9} & \textbf{83.5 ($\pm$14.0)} \\ \hline
VSE$\infty$ & LMH1                              & 80.8                           & \textbf{97.0} & \textbf{99.0} & 92.3 ($\pm$8.1)          &  & 66.0                           & 91.7                           & 96.1                           & 84.6 ($\pm$13.3)                           \\
            & LMH2                              & 81.0                           & 96.8                           & 99.0                           & 92.3 ($\pm$8.0)          &  & 66.4                           & \textbf{92.0}                           & 96.0                           & 84.8 ($\pm$13.1)                           \\
            & LSEH                              & \textbf{82.2} & 96.9                           & 98.6                           & \textbf{92.6 ($\pm$7.4)}          &  & \textbf{66.5} & 91.9 & \textbf{96.2} & \textbf{84.9 ($\pm$13.1)} \\ \hline
            
\end{tabular}
}
\end{table*}

\begin{table*}[htbp]
\caption{Results of cross-modal  information retrieval by LMH1, LMH2, and LSEH on the MS-COCO5K dataset in terms of average Recall@$k$ (\%).}
\label{tablePerformance3}
\centering
\resizebox{\linewidth}{!}{
\begin{tabular}{lccccrlcccr}
\hline
            & \multicolumn{1}{l}{}              & \multicolumn{4}{c}{Image-to-Text Retrieval}                                                                                 &  & \multicolumn{4}{c}{Text-to-Image Retrieval}                                                                                                   \\ \cline{3-6} \cline{8-11}
Network     & \multicolumn{1}{l}{Loss} & R@1                       & R@5                       & R@10                      & \multicolumn{1}{c}{Mean} &  & R@1                       & R@5                       & R@10                      & \multicolumn{1}{c}{Mean}                   \\ \hline
VSE++       & LMH1                              & 27.6                           & 56.6                           & 69.5                           & 51.2 ($\pm$17.5)         &  & \textbf{20.0} & 45.7                           & 59.0                           & 41.6 ($\pm$16.2)                           \\
            & LMH2                              & 3.0                            & 12.8                           & 20.9                           & 12.2 ($\pm$7.3)          &  & 0.4                            & 2.2                            & 3.7                            & 2.1 ($\pm$1.3)                             \\
            & LSEH                              & \textbf{28.5} & \textbf{57.6} & \textbf{70.3} & \textbf{52.1 ($\pm$17.5)}         &  & 19.3                           & \textbf{46.2} & \textbf{60.0} & \textbf{41.8 ($\pm$16.9)} \\ \hline
VSRN        & LMH1                              & 49.4                           & 79.3                           & 88.5                           & 72.4 ($\pm$16.7)         &  & 38.4                           & 69.3                           & 79.8                           & 62.5 ($\pm$17.6)                           \\
            & LMH2                              & 49.3                           & 79.4                           & 88.4                           & 72.4 ($\pm$16.7)         &  & 37.5                           & 68.3                           & 79.7                           & 61.8 ($\pm$17.8)                           \\
            & LSEH                              & \textbf{50.3} & \textbf{80.3} & \textbf{88.6} & \textbf{73.1 ($\pm$16.5)}         &  & \textbf{38.6} & \textbf{69.6} & \textbf{80.6} & \textbf{62.9 ($\pm$17.8)} \\ \hline
SGRAF-SAF   & LMH1                              & 54.4                           & 82.9                           & \textbf{91.0}                           & 76.1 ($\pm$15.7)         &  & 40.1                           & 69.7                           & 80.3                           & 63.4 ($\pm$17.0)                           \\
            & LMH2                              & 54.8                           & 82.8                           & 90.2                           & 75.9 ($\pm$15.2)         &  & 39.9                           & 69.3                           & 80.0                           & 63.1 ($\pm$17.0)                           \\
            & LSEH                              & \textbf{56.4} & \textbf{83.5} & 90.7 & \textbf{76.9 ($\pm$14.8)}         &  & \textbf{40.6} & \textbf{69.7} & \textbf{80.7} & \textbf{63.7 ($\pm$16.9)} \\ \hline
SGRAF-SGR   & LMH1                              & 56.8                           & 83.1                           & \textbf{91.1} & 77.0 ($\pm$14.7)         &  & \textbf{40.8} & 69.7                           & 80.6                           & 63.7 ($\pm$16.8)                           \\
            & LMH2                              & 55.9                           & 83.4                           & 90.8                           & 76.7 ($\pm$15.0)         &  & 39.6                           & 68.9                           & 79.7                           & 62.7 ($\pm$16.9)                           \\
            & LSEH                              & \textbf{57.8} & \textbf{83.7} & 91.0                           & \textbf{77.5 ($\pm$14.2)}         &  & 40.6                           & \textbf{69.8} & \textbf{80.6} & \textbf{63.7 ($\pm$16.9)} \\ \hline
VSE$\infty$ & LMH1                              & 58.3                           & 84.7                           & 91.8                           & 78.3 ($\pm$14.4)         &  & 42.5                           & 72.7                           & 83.0                           & 66.1 ($\pm$17.2)                           \\
            & LMH2                              & \textbf{58.9}                           & 84.9                           & 92.2                           & 78.7 ($\pm$14.3)         &  & 42.2                           & 72.4                           & 82.8                           & 65.8 ($\pm$17.2)                           \\
            & LSEH                              & 58.8 & \textbf{85.4} & \textbf{92.6} & \textbf{78.9 ($\pm$14.5)}         &  & \textbf{43.1} & \textbf{73.2} & \textbf{83.2} & \textbf{66.5 ($\pm$17.0)} \\ \hline
\end{tabular}
}
\end{table*}

\begin{table*}[htbp]
\caption{Results of cross-modal  information retrieval by LMH1, LMH2, and LSEH on the IAPR TC-12 dataset in terms of average Recall@$k$ (\%).}
\label{tablePerformance4}
\centering
\resizebox{\linewidth}{!}{
\begin{tabular}{lccccrlcccr}
\hline
            & \multicolumn{1}{l}{}              & \multicolumn{4}{c}{Image-to-Text Retrieval}                                                                                 &  & \multicolumn{4}{c}{Text-to-Image Retrieval}                                                                                                   \\ \cline{3-6} \cline{8-11}
Network     & \multicolumn{1}{l}{Loss} & R@1                       & R@5                       & R@10                      & \multicolumn{1}{c}{Mean} &  & R@1                       & R@5                       & R@10                      & \multicolumn{1}{c}{Mean}                   \\ \hline          
VSE++       & LMH1                              & 25.0                           & 56.5                           & 70.7                           & 50.7 ($\pm$19.1)         &  & 27.4                           & 60.9                           & 73.5                           & 53.9 ($\pm$19.5)                           \\
            & LMH2                              & 5.0                            & 14.2                           & 23.1                           & 14.1 ($\pm$7.4)          &  & 0.7                            & 3.5                            & 6.1                            & 3.4 ($\pm$2.2)                             \\
            & LSEH                              & \textbf{35.7} & \textbf{69.6} & \textbf{82.5} & \textbf{62.6 ($\pm$19.7)}         &  & \textbf{36.2} & \textbf{71.3} & \textbf{83.2} & \textbf{63.6 ($\pm$20.0)} \\ \hline
VSRN        & LMH1                              & 71.5                           & 92.9                           & 96.4                           & 86.9 ($\pm$11.0)         &  & 71.8                           & 92.8                           & 96.0                           & 86.9 ($\pm$10.7)                           \\
            & LMH2                              & 71.4                           & 93.4                           & \textbf{96.5} & 87.1 ($\pm$11.2)         &  & 70.8                           & \textbf{93.3}                           & \textbf{96.5}                          & 86.9 ($\pm$11.4)                           \\
            & LSEH                              & \textbf{73.9} & \textbf{93.8} & 96.2                           & \textbf{88.0 ($\pm$10.0)}         &  & \textbf{73.3} & 93.2 & 96.0 & \textbf{87.5 ($\pm$10.1)} \\ \hline
SGRAF-SAF   & LMH1                              & 70.7                           & 94.4                           & 97.4                           & 87.5 ($\pm$11.9)         &  & 73.1                           & 93.4                           & 97.2                           & 87.9 ($\pm$10.6)                           \\
            & LMH2                              & 9.8                            & 25.5                           & 37.1                           & 24.1 ($\pm$11.2)         &  & 1.9                            & 8.7                            & 14.0                           & 8.2 ($\pm$5.0)                             \\
            & LSEH                              & \textbf{74.5} & \textbf{95.1} & \textbf{97.8} & \textbf{89.1 ($\pm$10.4)}         &  & \textbf{73.7} & \textbf{94.5} & \textbf{97.7} & \textbf{88.6 ($\pm$10.6)} \\ \hline
SGRAF-SGR   & LMH1                              & 70.9                           & 93.3                           & 97.8                           & 87.3 ($\pm$11.8)         &  & 72.1                           & 92.8                           & 96.7                           & 87.2 ($\pm$10.8)                           \\
            & LMH2                              & 4.3                            & 12.7                           & 19.0                           & 12.0 ($\pm$6.0)          &  & 2.3                            & 9.7                            & 15.2                           & 9.1 ($\pm$5.3)                             \\
            & LSEH                              & \textbf{75.1} & \textbf{95.6} & \textbf{98.1} & \textbf{89.6 ($\pm$10.3)}         &  & \textbf{75.1} & \textbf{94.9} & \textbf{98.0} & \textbf{89.3 ($\pm$10.1)} \\ \hline
VSE$\infty$ & LMH1                              & 81.5                           & 97.3                           & \textbf{99.0} & 92.6 ($\pm$7.9)          &  & 79.1                           & 96.4                           & \textbf{98.8} & 91.4 ($\pm$8.8)                            \\
            & LMH2                              & 81.0                           & 96.7                           & 98.8                           & 92.2 ($\pm$7.9)          &  & 80.7                           & 95.8                           & 97.9                           & 91.5 ($\pm$7.7)                            \\
            & LSEH                              & \textbf{83.7} & \textbf{97.7} & 98.8                           & \textbf{93.4 ($\pm$6.9)}          &  & \textbf{81.4} & \textbf{96.9} & 98.5                           & \textbf{92.3 ($\pm$7.7)}  \\ \hline
\end{tabular}
}
\end{table*}

\begin{table*}[htbp]
\caption{Mean average Recall of five networks for each dataset. A summary derived from Table \ref{tablePerformance1}, \ref{tablePerformance2}, \ref{tablePerformance3}, and \ref{tablePerformance4}.}
\label{tablePerformanceAve}
\centering
\begin{tabular}{crrlrr}
\hline
\multicolumn{1}{l}{} & \multicolumn{2}{c}{Flickr30k} &  & \multicolumn{2}{c}{MS-COCO 1K} \\ \cline{2-3} \cline{5-6} 
\multicolumn{1}{l}{Method} & \multicolumn{1}{c}{Image-Text} & \multicolumn{1}{c}{Text-Image} &  & \multicolumn{1}{c}{Image-Text} & \multicolumn{1}{c}{Text-Image} \\ \hline
LMH1 & 82.9 ($\pm$15.6) & 71.0 ($\pm$17.3) &  & 88.8 ($\pm$10.9) & 80.3 ($\pm$16.3) \\
LMH2 & 65.6 ($\pm$35.1) & 56.3 ($\pm$31.3) &  & 79.7 ($\pm$25.2) & 68.3 ($\pm$32.6) \\
LSEH & \textbf{85.2 ($\pm$13.8)} & \textbf{73.0 ($\pm$16.6)} &  & \textbf{89.0 ($\pm$11.8)} & \textbf{80.5 ($\pm$16.4)} \\ \hline
\multicolumn{1}{l}{} & \multicolumn{2}{c}{MS-COCO 5K} & \multicolumn{1}{r}{} & \multicolumn{2}{c}{IAPR TC-12} \\ \cline{2-3} \cline{5-6} 
\multicolumn{1}{l}{Method} & \multicolumn{1}{c}{Image-Text} & \multicolumn{1}{c}{Text-Image} &  & \multicolumn{1}{c}{Image-Text} & \multicolumn{1}{c}{Text-Image} \\ \hline
LMH1 & 71.0 ($\pm$18.8) & 59.4 ($\pm$19.2) &  & 81.0 ($\pm$20.0) & 81.5 ($\pm$18.8) \\
LMH2 & 63.2 ($\pm$29.2) & 51.1 ($\pm$29.0) &  & 45.9 ($\pm$37.1) & 39.8 ($\pm$41.0) \\
LSEH & \textbf{71.7 ($\pm$18.5)} & \textbf{59.7 ($\pm$19.3)} &  & \textbf{84.5 ($\pm$16.5)} & \textbf{84.3 ($\pm$16.3)} \\ \hline
\end{tabular}
\end{table*}

\subsection{Comparison of LSEH with Polynomial Loss} \label{Comparison of LSEH with Polynomial Loss} 
This section compares the training efficiency and cross-modal information retrieval performance of the proposed LSEH with the LUWM polynomial loss function \cite{wei2021universal}. In Wei~et~al.~\cite{wei2021universal} LUWM was tested using GSMN, and hence the experiments included herein adopt the GSMN. This will facilitate the comparison of the performance of GSMN when using the LUWM and the proposed LSEH. 

\begin{figure}[ht]  
\centering 
\includegraphics[width=0.5\linewidth]{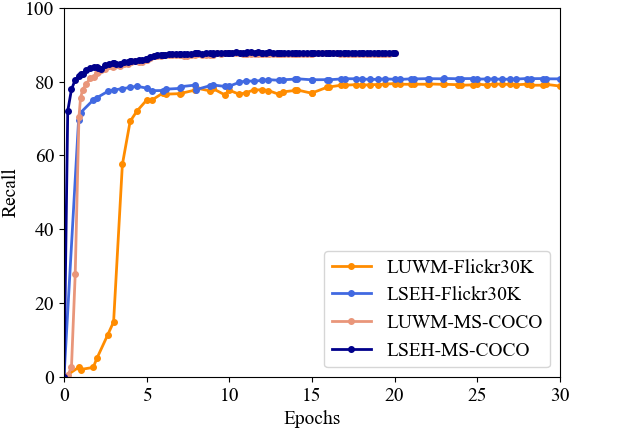} 
\caption{Comparison of training efficiency when using LUWM and LSEH to train GSMN on Flickr30K and MS-COCO respectively.} 
\label{figEfficiencyGSMN} 
\end{figure}

\textbf{Training Efficiency.} Fig. \ref{figEfficiencyGSMN} compares the training efficiency of GSMN using LUWM and LSEH on the Flickr30K and MS-COCO datasets, where the validation step for GSMN is at every 1000 mini-batches.  
Fig. \ref{figEfficiencyGSMN} shows that GSMN with LSEH achieved a larger M-Recall than with LUWM, at fewer epochs on both datasets. 

\textbf{Cross-modal Information Retrieval Performance.} 
Table \ref{tableRetrievalLUWMLSEH} shows the results of the comparisons of LUWM and LSEH when integrated into GSMN and applied to the datasets. The last column of Table \ref{tableRetrievalLUWMLSEH} shows the mean average Recall values (i.e. the average of columns 3-5), and the main findings from each dataset are summarised as follows. \\\textbf{(1) Flickr30K}. The proposed LSEH reached a Recall of 88.0\% for image-to-text, a Recall of 74.6\% for text-to-image retrieval, and outperformed LUWM by 0.5\% and 0.8\% for those tasks, respectively. \\\textbf{(2) MS-COCO 1K}. For image-to-text retrieval, LSEH's Recall reached 91.7\% which outperformed LMH1 by 1.2\%. For text-to-image retrieval, LSEH's Recall reached 83.0\% which outperformed LUWM by 1.2\%. \\\textbf{(3) MS-COCO 5K}. For LSEH, Recall reached 75.6\% for image-to-text and 62.3\% for text-to-image retrieval. LSEH outperformed LMH1 by 0.1\% and 0.6\% for image-to-text and text-to-image retrieval, respectively. 

\begin{table*}[!htbp]
\caption{Average Recall@$k$ (\%) of GSMN when using the LUWM and LSEH loss functions for cross-modal information retrieval on various datasets.}
\label{tableRetrievalLUWMLSEH}
\centering
\resizebox{0.9\linewidth}{!}{
\begin{tabular}{lccccrlcccr}
\hline
                      & \multicolumn{1}{l}{}              & \multicolumn{4}{c}{Image-to-Text Retrieval}                               &  & \multicolumn{4}{c}{Text-to-Image Retrieval}                               \\ \cline{3-6} \cline{8-11} 
Network               & \multicolumn{1}{l}{Loss} & R@1      & R@5      & R@10     & \multicolumn{1}{c}{Mean}  &  & R@1      & R@5      & R@10     & \multicolumn{1}{c}{Mean}  \\ \hline
\multicolumn{11}{c}{Flickr30K}                                                                                                                                                                                       \\ \hline
GSMN \cite{liu2020graph}     & LMH1                              & 71.4          & 92.0          & 96.1          & 86.5 ($\pm$10.8)          &  & 53.9          & 79.7          & 87.1          & 73.6 ($\pm$14.2)          \\
GSMN \cite{wei2021universal} & LUWM                              & 73.1          & 92.7          & \textbf{96.8} & 87.5 ($\pm$10.3)          &  & 54.2          & 79.9          & \textbf{87.3} & 73.8 ($\pm$14.2)          \\
GSMN                  & LSEH                              & \textbf{74.1} & \textbf{93.3} & 96.5          & \textbf{88.0 ($\pm$9.9)}  &  & \textbf{55.4} & \textbf{81.2} & 87.2          & \textbf{74.6 ($\pm$13.8)} \\ \hline
\multicolumn{11}{c}{MS-COCO 1K}                                                                                                                                                                                      \\ \hline
GSMN \cite{liu2020graph}     & LMH1                              & 76.1          & 95.6          & 98.3          & 90.0 ($\pm$9.9)           &  & 60.4          & 88.7          & 95.0          & 81.4 ($\pm$15.0)          \\
GSMN \cite{wei2021universal} & LUWM                              & 76.8          & 96.2          & 98.5          & 90.5 ($\pm$9.7)           &  & 60.9          & 89.0          & \textbf{95.5} & 81.8 ($\pm$15.0)          \\
GSMN                  & LSEH                              & \textbf{79.7} & \textbf{96.4} & \textbf{98.9} & \textbf{91.7 ($\pm$8.5)}  &  & \textbf{63.2} & \textbf{90.3} & 95.4          & \textbf{83.0 ($\pm$14.1)} \\ \hline
\multicolumn{11}{c}{MS-COCO 5K}                                                                                                                                                                                      \\ \hline
GSMN \cite{wei2021universal} & LUWM                              & \textbf{54.7} & 82.2          & 89.7          & 75.5 ($\pm$15.0)          &  & 38.5          & 67.6          & 78.9          & 61.7 ($\pm$17.0)          \\
GSMN                  & LSEH                              & 54.3          & \textbf{82.4} & \textbf{90.2} & \textbf{75.6 ($\pm$15.4)} &  & \textbf{39.0} & \textbf{68.4} & \textbf{79.5} & \textbf{62.3 ($\pm$17.1)} \\ \hline
\end{tabular}
 }
\end{table*}

\subsection{Discussion on Quantitative Results of Graph-based VSE Networks} 
Graph-based VSE networks such as SGRAF-SGR, VSRN, and GSMN consider the local information of images and descriptions for cross-modal information retrieval. The results described in Section \ref{ResultsEfficiency}-\ref{Comparison of LSEH with Polynomial Loss} revealed that the above graph-based VSE networks performed better than when using the LMH and LUWM loss functions. 
When using LMH1, SGRAF-SGR needed 38.6 training epochs to achieve its higher M-Recall value (81.0\%) on the Flickr30K dataset, compared to needing 20.1 epochs when using LSEH. 
Hence when using LSEH the number of training epochs needed by SGRAF-SGR were reduced by 47.9\%. 
In terms of retrieval performance using the mean average Recall evaluation metric, (1) VSRN using LSEH reached the Recall of 87.2\% and 75.5\% for image-to-text and text-to-image retrieval respectively on the Flickr30K dataset, and outperformed VSRN when using LMH1 by 2.8\% and 3.0\% for those tasks, respectively; 
(2) GSMN using LSEH reached a Recall of 91.7\% for image-to-text, and a Recall of 83.0\% for text-to-image retrieval on the MS-COCO1K dataset, and outperformed GSMN when using LUWM by 1.2\% and 1.2\% for those tasks, respectively.

\subsection{Comparisons on Computation Time} \label{Comparisons on Computation Time} 
\begin{figure}[htbp]
\centering
    \includegraphics[width=0.5\linewidth]{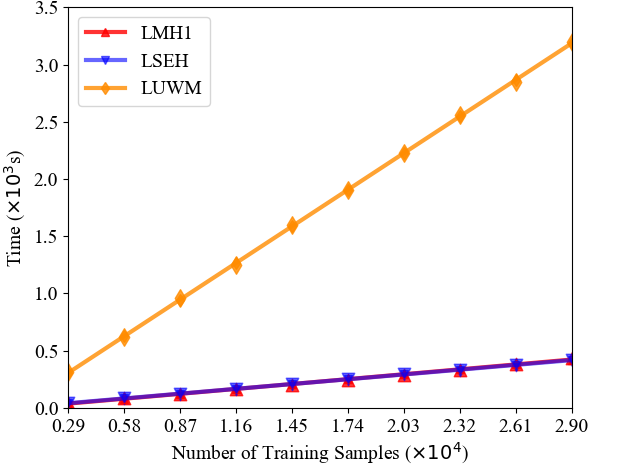}
	\caption{Computation time at 1 epoch when using LMH1, LUWM, and LSEH to train VSE$\infty$.}
	\label{TrainingTime}
\end{figure}

Fig.~\ref{TrainingTime} shows the computation time of one epoch when training VSE$\infty$ \cite{chen2021learning} using LMH1, LUWM, and LSEH. These experiments utilised the training set from Flickr30K \cite{young2014image}. In Fig. \ref{TrainingTime}, for each loss function the network is trained 10 times with various numbers of training samples (increased from 2900 to 29000 in steps of 2900), and the computation time is recorded. The three lines that fit the data points of LMH1, LUWM, and LSEH follow the equations of $T_{LMH1} (n) = 147.8n - 6.4$, $T_{LUWM} (n) = 1104.6n-14.7$, and $T_{LSEH} (n) = 144.4n - 1.7$ respectively. Note that the lines of LSEH and LMH1 are almost aligned. The computation time of LSEH is almost the same as that of LMH1, and six times faster than LUWM. 

\subsection{Comparison of LSEH-SVD and LSEH-BERT} \label{Comparing SVD and BERT} 
This section compares LSEH when using SVD (LSEH-SVD) and when using BERT (LSEH-BERT) for cross-modal information retrieval in the Flickr30K, MS-COCO, and IAPR TC-12 benchmark datasets. The experiments employ the latest VSE network (i.e. VSE$\infty$) to test LSEH-SVD and LSEH-BERT. VSE$\infty$ using LSEH-BERT employs the pre-trained BERT-base model \cite{devlin2019bert} and follows the hyperparameter settings of VSE$\infty$ using LSEH-SVD (shown in Table \ref{Hyperparameters}). 

\begin{figure}[htbp]
\centering
\includegraphics[width=0.5\linewidth]{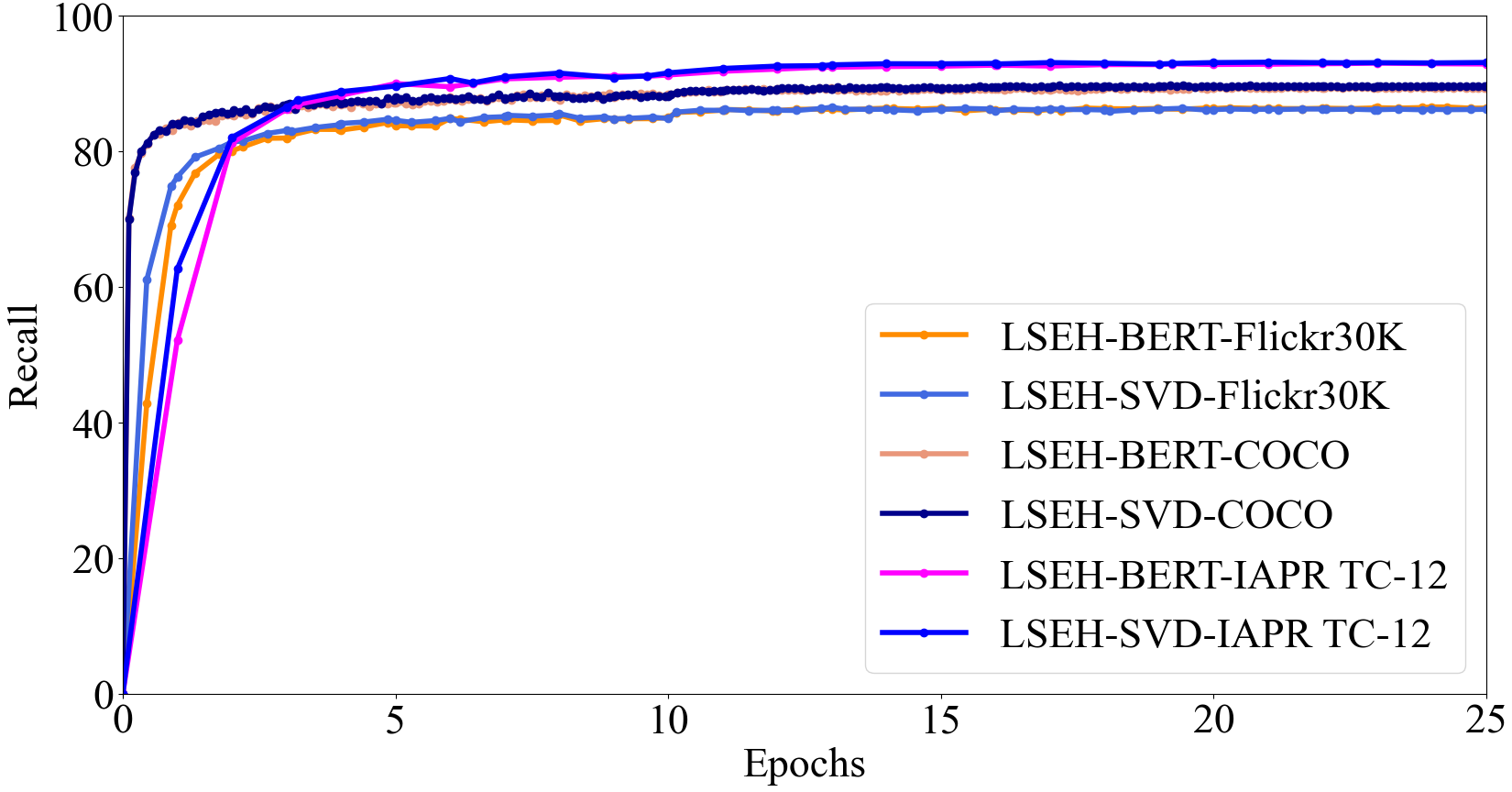}
\caption{Comparison of training efficiency when using LSEH-SVD and LSEH-BERT to train VSE$\infty$ on the Flickr30K, MS-COCO, and IAPR TC-12 datasets.}
\label{figEfficiencyBERT}
\end{figure}

\textbf{Training Efficiency.} Fig. \ref{figEfficiencyBERT} compares the training efficiency of VSE$\infty$ using LSEH-SVD and LSEH-BERT, where the validation step for VSE$\infty$ is at every 500 mini-batches. Fig. \ref{figEfficiencyBERT} shows that VSE$\infty$ with LSEH-SVD reached a larger M-Recall than with LSEH-BERT at fewer epochs on the Flickr30K and IAPR TC-12 datasets. Observing the curves of VSE$\infty$ using LSEH-SVD and LSEH-BERT on the MS-COCO dataset, their lines are close to overlapping and hence they are similar with regards to M-Recall and training efficiency. 

\textbf{Cross-modal Information Retrieval Performance.} As shown in Table \ref{tableRetrievalSVDBERT} , compared to the baseline (LMH1), both LSEH-SVD and LSEH-BERT have outperformed LMH1 for cross-modal information retrieval on all datasets. The performance of LSEH-BERT is lower than that of LSEH-SVD. LSEH-BERT outperformed LSEH-SVD only for the task of image-to-text retrieval on the dataset of MS-COCO 5K, where LSEH-BERT reached a mean average Recall value of 79.0\% and outperformed LSEH-SVD by 0.1\%. However, LSEH-SVD outperformed LSEH-BERT on all the rest of the tasks and datasets in Table \ref{tableRetrievalSVDBERT}. 

\begin{table*}[!htbp]
\caption{Average Recall@$k$ (\%) of VSE$\infty$ when using the LSEH-SVD and LSEH-BERT for cross-modal information retrieval on various datasets.}
\label{tableRetrievalSVDBERT}
\centering
\resizebox{0.9\linewidth}{!}{
\begin{tabular}{lccccrlcccr}
\hline
              &        & \multicolumn{4}{c}{Image-to-Text Retrieval}                               &  & \multicolumn{4}{c}{Text-to-Image Retrieval}                               \\ \cline{3-6} \cline{8-11} 
Loss & Method & R@1      & R@5      & R@10     & \multicolumn{1}{c}{Mean}  &  & R@1      & R@5      & R@10     & \multicolumn{1}{c}{Mean}  \\ \hline
\multicolumn{11}{c}{Flickr30K}                                                                                                                                                    \\ \hline
LMH1          & -      & 80.8          & \textbf{96.4} & 98.3          & 91.8 ($\pm$7.8)           &  & 62.6          & 86.9          & 91.7          & 80.4 ($\pm$12.7)          \\
LSEH          & BERT   & 81.0          & 96.1          & 97.7          & 91.6 ($\pm$7.5)           &  & 63.1          & 87.1          & \textbf{92.7} & 81.0 ($\pm$12.8)          \\
LSEH          & SVD    & \textbf{82.4} & 96.0          & \textbf{98.6} & \textbf{92.3 ($\pm$7.1)}  &  & \textbf{63.7} & \textbf{87.1} & 92.5          & \textbf{81.1 ($\pm$12.5)} \\ \hline
\multicolumn{11}{c}{MS-COCO 1K}                                                                                                                                                   \\ \hline
LMH1          & -      & 80.8          & 97.0          & \textbf{99.0}          & 92.3 ($\pm$8.1)           &  & 66.0          & 91.7          & 96.1          & 84.6 ($\pm$13.3)          \\
LSEH          & BERT   & 82.0          & \textbf{97.1} & 98.7 & 92.6 ($\pm$7.5)           &  & 66.2          & \textbf{92.0} & 96.2          & 84.8 ($\pm$13.3)          \\
LSEH          & SVD    & \textbf{82.2} & 96.9          & 98.6          & \textbf{92.6 ($\pm$8.5)}  &  & \textbf{66.5} & 91.9          & \textbf{96.2} & \textbf{84.9 ($\pm$13.1)} \\ \hline
\multicolumn{11}{c}{MS-COCO 5K}                                                                                                                                                   \\ \hline
LMH1          & -      & 58.3          & 84.7          & 91.8          & 78.3 ($\pm$14.4)          &  & 42.5          & 72.7          & 83.0          & 66.1 ($\pm$17.2)          \\
LSEH          & BERT   & \textbf{59.1} & 85.3          & 92.6          & \textbf{79.0 ($\pm$14.4)} &  & 42.0          & 72.7          & 83.1          & 65.9 ($\pm$17.4)          \\
LSEH          & SVD    & 58.8          & \textbf{85.4} & \textbf{92.6} & 78.9 ($\pm$14.5)          &  & \textbf{43.1} & \textbf{73.2} & \textbf{83.2} & \textbf{66.5($\pm$17.0)}  \\ \hline
\multicolumn{11}{c}{IAPR TC-12}                                                                                                                                                   \\ \hline
LMH1          & -      & 81.5          & 97.3          & \textbf{99.0} & 92.6 ($\pm$7.9)           &  & 79.1          & 96.4          & \textbf{98.8} & 91.4 ($\pm$8.8)           \\
LSEH          & BERT   & 82.9          & 97.4          & 98.5          & 92.9 ($\pm$7.1)           &  & 80.6          & 96.5          & 97.9          & 91.7 ($\pm$7.8)           \\
LSEH          & SVD    & \textbf{83.7} & \textbf{97.7} & 98.8          & \textbf{93.4 ($\pm$6.9)}  &  & \textbf{81.4} & \textbf{96.9} & 98.5          & \textbf{92.3 ($\pm$7.7)}  \\ \hline
\end{tabular}
 }
\end{table*}

\subsection{Reasoning with the Results of the Proposed LSEH using Visualisation} 
\begin{figure*}[ht] 
\centering 
\includegraphics[width=\linewidth]{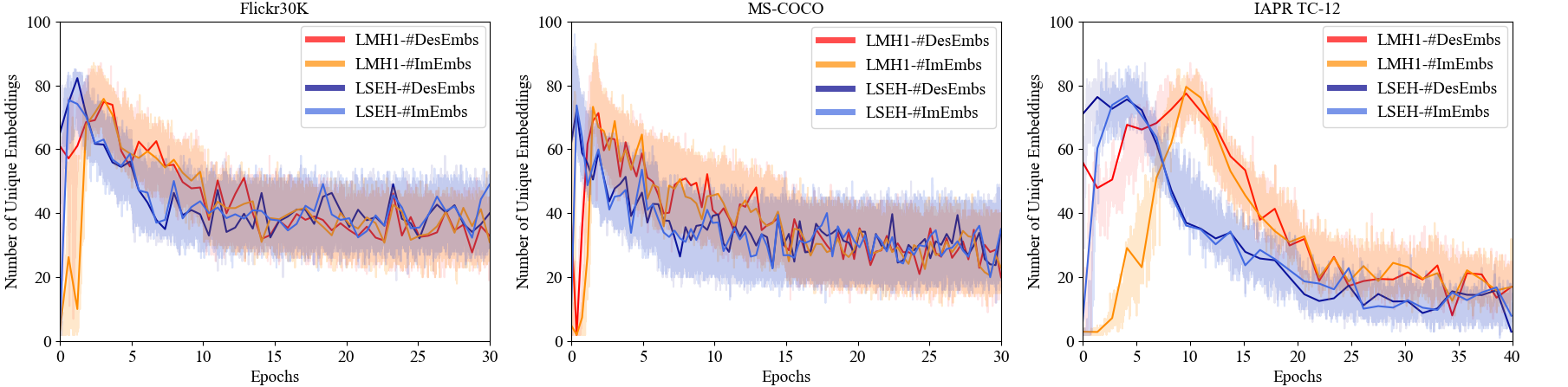} 
\caption{Results of reasoning with training efficiency of VSRN using LMH1 and LSEH on various datasets. The Y-axis presents the numbers of unique image embeddings (\#ImEmbs) and description embeddings (\#DesEmbs) that are used as hard negative samples in each mini-batch, and the X-axis is the number of training epochs.} 
\label{whyfaster} 
\end{figure*} 

\begin{figure*}[ht] 
\centering 
\includegraphics[width=0.9\linewidth]{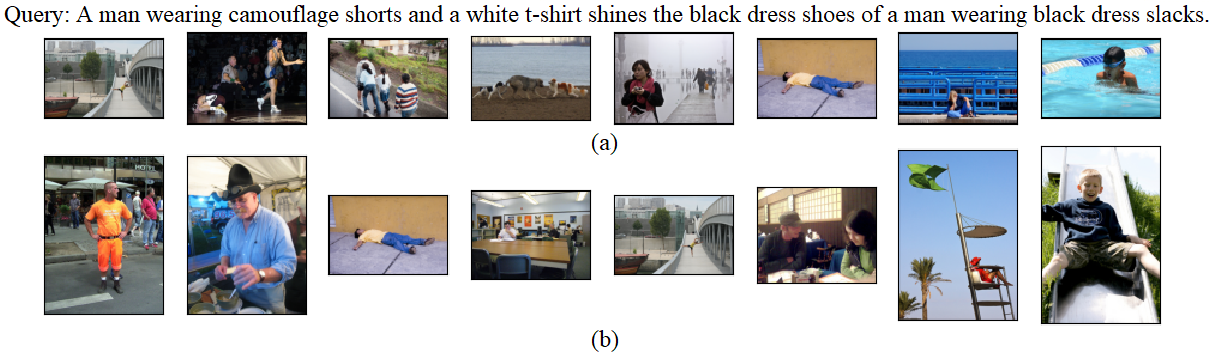} 
\caption{Examples (a) and (b) are the top eight irrelevant images in a mini-batch to the query description that is selected by LMH1 and LSEH respectively according to the descending order of their gradients. The images in (b) are semantically closer to the query’s textual description than the images in (a), which are hard negative samples.} 
\label{expHN} 
\end{figure*} 

\begin{figure*}[htbp] 
\centering 
\includegraphics[width=0.8\linewidth]{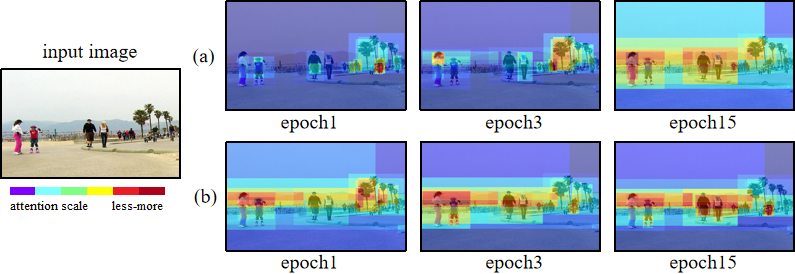} 
\caption{Visual comparison of LMH1 (a) and LSEH (b) on the various attention regions of an image processed by VSRN during training. The degree of attention is indicated on the attention scale.} 
\label{visualizationE1} 
\end{figure*} 
For reasoning with how the proposed LSEH improved the training of VSE networks, VSRN \cite{li2022image} was adopted because it is a state-of-the-art image graph reasoning network that provides the means of visualising its results.  

\textbf{Reasoning with Training Efficiency.} When the queries are looped in a mini-batch during training, LMH1 and LSEH select one irrelevant embedding from a set of irrelevant embeddings as the hard negative sample for the query embedding, because of the max operation in LMH1 and LSEH (as shown in Eq. (\ref{equationLMH}) and Eq. (\ref{equationLSEH})). However, some irrelevant embeddings may be repeatedly selected as the hard negative samples and this leads to low efficiency when utilising embeddings (see Section \ref{Proposed Semantically-Enhanced Hard Negatives Loss Function}). In this experiment, to illustrate and compare the training efficiency of VSRN when using LMH1 and the proposed LSEH loss function, the number of unique image-- and description-- embeddings (denoted as \#ImEmbs and \#DesEmbs respectively), that are used as the hard negative samples, are counted within each mini-batch and across a range of training epochs.  

The results of each dataset are shown in Fig. \ref{whyfaster} and described as follows. \\\textbf{(1) Flickr30K}. LSEH reached its largest number of unique \#DesEmbs and \#ImEmbs (i.e. both 85) at 1.1 epochs, whereas LMH1 needed 2.9 epochs (i.e. \#DesEmbs and \#ImEmbs were both 87). \\\textbf{(2) MS-COCO}. LSEH needed 0.2 epochs to reach its largest number of \#DesEmbs (91) and \#ImEmbs (i.e. 96), whereas LMH1 needed 1.7 epochs to obtain its largest number of unique \#DesEmbs and \#ImEmbs with values 89 and 93 respectively. \\\textbf{(3) IAPR TC-12}. LSEH reached its largest number of unique \#DesEmbs (i.e. 88) and \#ImEmbs (i.e. 87) in 2.2 epochs, whereas LMH1 needed 10 epochs to obtain its largest number of unique \#DesEmbs (i.e 87) and \#ImEmbs (i.e 85). The proposed LSEH outperformed LMH1 on the efficiency of utilising unique embeddings, which visually reasons the improvement in training efficiency by LSEH. 

\textbf{Illustrating one Hard Negative Sample Selected by LMH1 and LSEH.} For visually comparing LMH1 and LSEH when selecting hard negative samples, the irrelevant images to each textual query found in a mini-batch are ranked by LMH1 and LSEH respectively according to the descending order of their gradients. The top eight irrelevant images to a sample textual query are presented in Fig. \ref{expHN}. Images selected by LMH1 in Fig. \ref{expHN} (a) contain no semantic similarity relevant to the query description. However, in comparison to the set of results retrieved by LSEH and shown in set (b) of Figure \ref{expHN}, the selected images are more semantically relevant to the query, e.g. more images were retrieved of `a man' or which relate to the clothes described in the query. The ranked top eight irrelevant images selected by LSEH have a higher probability of being incorrectly retrieved than those irrelevant images selected by LMH1, hence why it is more efficient to use LSEH's hard negative samples during the training.

\textbf{Visualisation of Learning with LMH1 and LSEH.} For visually comparing the learning progress of VSRN when using LMH1 and LSEH, the various attention regions of an image processed by VSRN during the training are presented, where the method for generating the attention region heating maps follows that of Li~et~al.~\cite{li2022image}. VSRN using LMH1, as shown in Fig. \ref{visualizationE1} (a) only focused some attention on the main objects of the image (e.g. `people' and `trees') on epochs one and three. On the contrary, as shown in Fig. \ref{visualizationE1} (b), the VSRN using the proposed LSEH has paid much attention to the main objects of the image on epoch one. 

\section{Conclusion}
Existing Visual Semantic Embedding (VSE) networks are trained by a hard negatives loss function that learns an objective margin between the similarity of the relevant and irrelevant image--description embedding pairs and ignores the semantic differences between the irrelevant pairs. This paper proposes a novel Semantically-Enhanced Hard negatives Loss function (LSEH) for Cross-modal Information Retrieval that considers the semantic differences between irrelevant training pairs, to dynamically adjust the learning objectives of VSE networks to make their learning flexible and efficient. 
Extensive experiments were carried out by integrating the proposed methods into five state-of-the-art VSE networks (VSE++, VSRN, SGRAF-SAF, SGRAF-SGR, and VSE$\infty$) that were applied to the Flickr30K, MS-COCO, and IAPR TC-12 datasets. The experiments revealed that the proposed LSEH function, when integrated into VSE networks, improves their training efficiency and cross-modal information retrieval performance. 
With regards to training time, LSEH reduced the training epochs of LMH on average by 53.2\% on the Flickr30K dataset, 43.0\% on the MS-COCO dataset, and 48.5\% on the IAPR TC-12 dataset. In terms of retrieval performance using the mean average Recall evaluation metric, LSEH outperformed LMH1 by (1) 2.3\% for image-to-text and 2.0\% for text-to-image retrieval on the Flickr30K dataset; (2) 0.2\% for image-to-text and 0.2\% for text-to-image retrieval on the MS-COCO1K dataset; (3) 0.7\% for image-to-text and 0.3\% for text-to-image retrieval on the MS-COCO5K dataset; and (4) 3.5\% for image-to-text and 2.8\% for text-to-image retrieval on the IAPR TC-12 dataset. 
Section \ref{Comparing SVD and BERT} describes our experimental results using SVD and BERT and the findings revealed that SVD outperformed BERT when these are embedded in the proposed loss function. From this perspective, it is more efficient to use an unsupervised approach because such approaches don't need labels or computationally expensive training processes. 
The focus of the paper has been on the application of VSE networks for cross-modal information retrieval. Future work includes applying the proposed loss function to other similar cross-modal tasks such as hashing-based networks for cross-modal retrieval and text-to-video retrieval. 
Future work will also focus on extracting image description semantics using other supervised and unsupervised feature extraction methods and comparing those to the methods (i.e. SVD and BERT) that have already been utilised for the task. Finally, future work also includes embedding the proposed methods into a next generation cross-modal search engine and evaluating its capabilities with real-word datasets. 
\bibliographystyle{unsrt}  
\bibliography{references}  

\end{document}